\documentclass{article}
\usepackage{iclr2027_conference,times}

\usepackage{amsmath}
\usepackage{amssymb}
\usepackage{amsfonts}
\usepackage{bm}
\usepackage{graphicx}
\usepackage{fontawesome5}
\usepackage{wrapfig}
\usepackage{booktabs}
\usepackage{multirow}
\usepackage{xcolor}
\usepackage{colortbl}
\usepackage{array}
\usepackage{microtype}
\usepackage{hyperref}
\usepackage{titletoc}
\usepackage{url}
\usepackage{enumitem}
\usepackage{pifont}
\usepackage{float}

\newcommand{\sigv}{\sigma_{v}}
\newcommand{\siga}{\sigma_{a}}
\newcommand{\sigi}{\sigma_{i}}

\newcommand{\vz}{\bm{v}}
\newcommand{\az}{\bm{a}}
\newcommand{\xz}{\bm{x}}
\newcommand{\refv}{\tilde{\bm{v}}}
\newcommand{\refa}{\tilde{\bm{a}}}
\newcommand{\refx}{\tilde{\bm{x}}}
\newcommand{\model}{UniAVC}
\newcommand{\ie}{\emph{i.e.}}

\newcommand{\E}[2]{\mathbb{E}_{#1}\!\left[#2\right]}

\newcommand{\ours}{\rowcolor{gray!15}}

\usepackage{color}
\definecolor{citecolor}{RGB}{66,168,235}
\definecolor{linkcolor}{RGB}{255,0,0}

\hypersetup{colorlinks=true,citecolor=citecolor,linkcolor=linkcolor}

\title{\model: Unified Audio-Video Creation via Progressive Training and Asymmetric Scheduling}

\iclrfinalcopy 
\author{%
  Kien T. Pham$^{1}$, Yuhao Liu$^{2, \ddagger}$, Chong Hou Choi$^{2}$, Zhenqi He$^{1}$, Yanghao Wang$^{1}$, \\
  \textbf{Chenyang Lei$^{2}$, Rui Liu$^{2,\dagger}$, Qifeng Chen$^{1,\dagger}$, Long Chen$^{1,\dagger}$} \\[2pt]
  $^{1}$HKUST \qquad $^{2}$Huawei Research\\
  \texttt{tkpham@connect.ust.hk}, \texttt{longchen@ust.hk}
}

\begin{document}

\maketitle
\lhead{Preprint}

{
  \renewcommand{\thefootnote}{\fnsymbol{footnote}}
  \footnotetext[3]{Project leader.}
  \footnotetext[2]{Co-corresponding authors.}
}

\vspace{-0.25in}
{\centering\small\bfseries
  \definecolor{resourceblue}{HTML}{1772D0}%
  \hypersetup{urlcolor=resourceblue}%
  \mbox{\href{https://uniavc.tkpham.me}{\textcolor[HTML]{147D92}{\faGlobe}\enspace Project Page}}\qquad
  \mbox{\href{https://github.com/HKUST-LongGroup/uniavc-model}{\textcolor{black}{\faGithub}\enspace Code}}\qquad
  \mbox{\href{https://huggingface.co/TK3105/UniAVC}{\raisebox{-0.15em}{\includegraphics[height=1.15em]{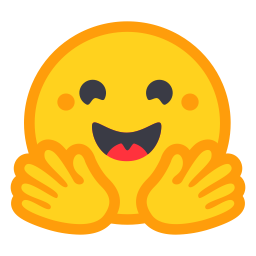}}\enspace Model}}\par}
\vspace{0.1in}

\begin{abstract}
    Unified audio-video creation requires a single model capable of text-conditioned joint synthesis, bidirectional cross-modal generation, and instruction-guided editing across both modalities. Existing methods rely heavily on paired corpora, which are scarce for joint generation and even rarer for editing. Consequently, they are spent learning single-branch generative and editing capabilities that abundant unimodal data could readily teach, rather than on the cross-modal coupling. Furthermore, prevailing systems apply a single shared noise schedule to both modalities, neglecting two critical limitations. First, due to the significant discrepancy in information density, video requires prolonged high noise to organize spatial and motion structure, whereas the sparser audio signal can settle earlier during the synthesis process, the dynamics a uniform schedule might compromise. Second, lockstep denoising leaves asymmetric workflows, such as conditioning on one modality to synthesize or edit another, untrained. We present \textbf{\model}, which addresses both obstacles. It first trains each branch on unimodal dataset to full generative and editing competence, and subsequently dedicates paired data to exploit cross-modal correspondence. Modality-decoupled schedulers meanwhile warp the noise schedule independently from a shared timestep, while subtasks with one clean modality directly supervise asymmetric configurations. These settings respectively enable intent-aware editing, which preserves one modality intact while enlisting it to constrain the other, and post-inference self-refinement, which regenerates only an unsatisfactory branch against an accepted one. \model{} thus spans multiple creation tasks from a single set of weights without task-specific adapters. Extensive experiments demonstrate that \model{} achieves versatile audio-video generation and editing with competitive performance across different operating modes.
    \end{abstract}
    
\section{Introduction}
\label{sec:intro}

Audio-video content creation has advanced along several largely independent lines. Joint generators
synthesize a clip together with its soundtrack in a single pass, achieving a synchronization that
cascaded pipelines struggle to match \citep{ovi,universe1,pham2026avtok,ltx2,jova}. Cross-modal models generate one modality
conditioned on the other \citep{luo2023difffoley,zhang2024foleycrafter,cheng2024mmaudio}.
Instruction-guided editors, extending a paradigm established for images
\citep{brooks2023instructpix2pix,zhang2024magicbrush}, modify an existing clip according to a
textual request \citep{javedit,instructav2av}, and control adapters attach conditioning to a pretrained
backbone, typically one adapter per control \citep{avcontrol}. Each line has matured on its own
terms, and each ships its own weights, training corpus, and task coverage.

\begin{figure}[t]
\begin{center}
\includegraphics[width=\textwidth]{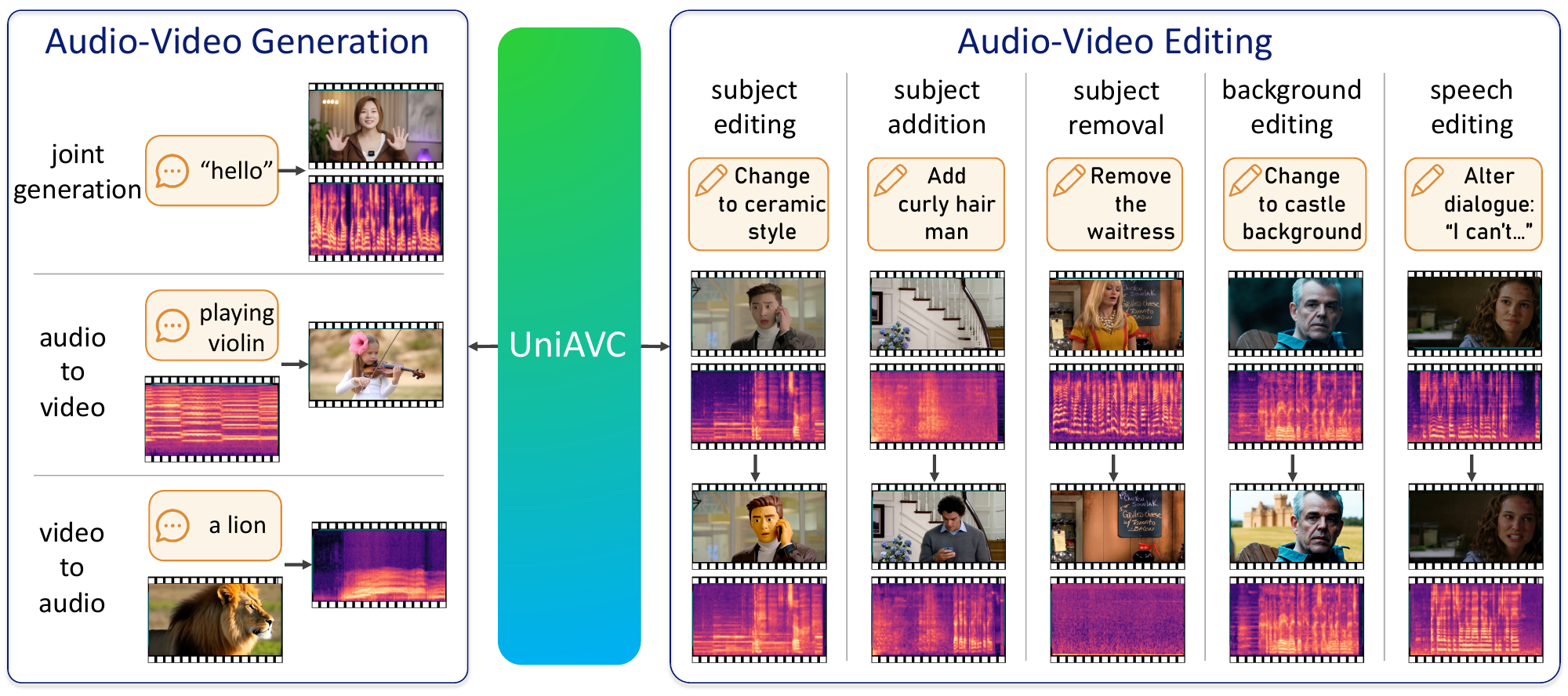}
\end{center}
\caption{\model{} performs joint audio-video generation, cross-modal generation in either direction,
and instruction-guided editing of either or both modalities from one set of weights.}
\label{fig:teaser}
\vspace{-2.em}
\end{figure}


Operating across these decoupled paradigms carries a severe cost for audio-video creation. 
Because existing tools specialize in either generation or editing, practical workflows must cascade independent models sequentially, for instance, by generating a synchronized clip with a joint model and then modifying it with another instruction-guided editor. 
This sequential pipeline approach inherently degrades cross-modal coherence. 
When an editor modifies the visual stream without awareness of the soundtrack, it immediately breaks the precise synchrony established during initial generation. 
This drift accumulates through iterative edits, such as modifying ambient noise or restyling a visual scene, forcing users to deploy dedicated resynchronization modules or explicit alignment losses just to repair the audio-visual match \citep{javedit,spongebob}. 
A single model that natively unifies generation and editing across both modalities eliminates this vulnerability, establishing temporal correspondence once and retaining it across all downstream revisions \citep{jova}. 
Realizing such a unified framework, however, remains markedly harder than assembling specialized tools due to two compounding challenges.

\textbf{C1: Scarcity of paired audio-video supervision.}
High-quality datasets offering both precise audio-visual alignment for generation and instruction-guided editing pairs are far smaller than massive unimodal corpora. 
Training a unified model directly on this limited paired data yields poor convergence, as the network must simultaneously master multiple competing objectives, including source preservation, text instruction compliance, and cross-modal synchronization \citep{instructav2av}. 
Exacerbating this bottleneck, much of this scarce paired supervision is inefficiently spent teaching the model basic single-modality capabilities, such as rendering realistic visual motion or synthesizing clear acoustic signals. 
Because unimodal datasets already provide these isolated features in abundance, relying on paired data to learn them dilutes supervision signal through redundant learning, leaving inadequate learning capacity to capture the intricate cross-modal coupling that unimodal data alone can never supply.

\begin{wrapfigure}{r}{0.42\linewidth}
\vspace{-10pt}
\centering
\includegraphics[width=.95\linewidth]{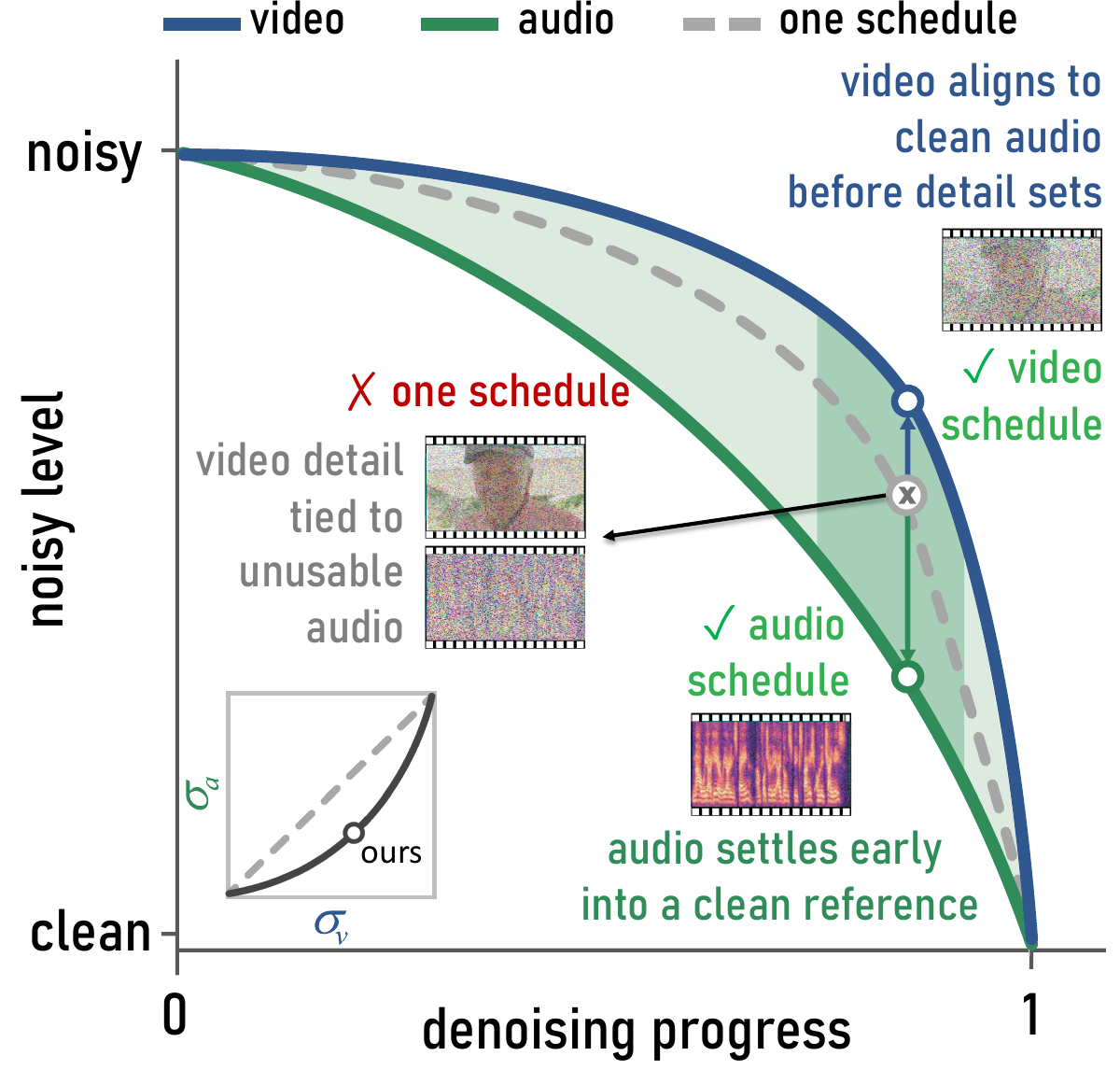}
\vspace{-1.3em}
\caption{Single vs.\ dual noise schedules.}
\label{fig:c2}
\end{wrapfigure}
\textbf{C2: Neglected information asymmetry between audio and video representations.}
Video latents maintain a dense spatio-temporal structure across every frame \citep{blattmann2023stable,hacohen2024ltxvideo}, whereas audio codecs compress signals into a far sparser sequence \citep{liu2023audioldm,evans2024stableaudio}. 
Consequently, a clip's visual token stream outnumbers its acoustic counterpart by more than an order of magnitude \citep{ltx2}. 
Prevailing joint architectures nevertheless corrupt both modalities along a single shared noise schedule, leaving this structural disparity unaddressed. 
Because video requires a prolonged high-noise phase to establish global texture, identity, and motion before refining fine details, its optimal denoising trajectory differs fundamentally from that of audio, which settles into its final acoustic content much earlier (Fig.~\ref{fig:c2}). 
Aligning the shared schedule to video leaves audio artificially noisy long after it has converged, depriving the visual stream of clean cross-modal conditioning. 
Conversely, optimizing the schedule for audio forces video to commit to details prematurely before its global structure has formed. 
Ultimately, a single rigid schedule fails both modalities.
Beyond ignoring these distinct denoising trajectories, enforcing a shared schedule severely restricts how the model explores noise states during training. 
Because each modality carries an independent noise level, the true state space during training is two-dimensional \citep{bao2023unidiffuser}, rather than the single scalar timeline assumed in standard diffusion frameworks \citep{ho2020ddpm,song2021scoresde}. 
Locking both streams to equal noise levels forces training strictly along the diagonal of this 2D space, teaching the model only how to denoise symmetrically corrupted pairs. The core capabilities promised by a unified framework, however, depend on off-diagonal, asymmetric noise states. 
Tasks such as restyling a visual scene while preserving its soundtrack or generating video conditioned on existing audio require denoising a corrupted video stream against completely clean audio. 
Because a shared schedule never exposes the network to these off-diagonal configurations during training, the model encounters a severe distribution shift at inference, asking it to execute conditioning behaviors it was never trained to perform.

\textbf{Our approach: \model{}.} To address these challenges, we introduce \model{}, a unified audio-video creation model built with progressive training and decoupled noise scheduling. 
We first establish full generative and editing competence in each modality using abundant unimodal corpora, then freeze both branches so that scarce paired data is used exclusively to train lightweight cross-modal bridges. 
This guarantees that paired supervision is dedicated entirely to cross-modal coupling rather than redundant single-modality learning. 
To address representation asymmetry and explore off-diagonal noise space, asymmetric modality-decoupled schedulers map a shared timestep index onto two independently warped noise trajectories, curving the training region away from the symmetric diagonal. 
Complementing this, clean-branch subtasks explicitly supervise extreme boundary configurations that continuous warped curves cannot reach. 
This joint coverage unlocks intent-aware editing, which holds the preserved modality as a clean reference to return the source material unchanged, as well as post-inference self-refinement, enabling the selective regeneration of an unsatisfactory branch conditioned on an accepted one.


\textbf{Our primary contributions are summarized as follows:}
\begin{itemize}[leftmargin=*, noitemsep]
\item \textbf{Unified generation and editing framework.} A single model natively performs text-conditioned joint audio-video generation, bidirectional cross-modal synthesis, and instruction-guided editing of either or both modalities without requiring task-specific adapters (Fig.\ref{fig:teaser}).
\item \textbf{Progressive training paradigm.} By first mastering unimodal generative and editing capabilities on single-modality corpora before freezing both branches, our framework dedicates scarce paired datasets entirely to establishing cross-modal correspondence and temporal synchronization.
\item \textbf{Asymmetric noise-level scheduling with boundary supervision.} Modality-decoupled schedulers derive per-branch noise trajectories from a shared timestep index to steer training off the diagonal of the 2D noise space, while clean-branch subtasks explicitly supervise extreme boundary states to enable intent-aware editing and post-inference self-refinement.
\item \textbf{Extensive empirical evaluation.} Comprehensive experiments across diverse generation and editing benchmarks demonstrate that \model{} achieves performance competitive with specialized state-of-the-art baselines across all supported creation modes.
\end{itemize}
\section{Related Work}
\label{sec:related}

\textbf{Joint Audio-Video Generation. } Early joint synthesis coupled distinct modality-specific diffusion processes, either through shared cross-modal architectures \citep{ruan2023mmdiffusion} or by aligning independently trained generators during inference \citep{xing2024seeingandhearing}. In parallel, cross-modal models synthesize one stream conditioned on the other, deriving alignment directly from the conditioning input rather than learning a joint prior \citep{luo2023difffoley,zhang2024foleycrafter,cheng2024mmaudio,pham2025spa2v}. Contemporary joint frameworks scale these paradigms by fusing twin backbones block by block \citep{ovi}, integrating specialized experts into a shared trunk \citep{universe1,mova}, estimating hierarchical spatio-temporal priors before generation \citep{javisdit,javisditpp}, or training end-to-end audio-visual foundation models \citep{ltx2}. These advances leverage video diffusion transformers \citep{blattmann2023stable,hacohen2024ltxvideo,kong2024hunyuanvideo,polyak2024moviegen} and audio latent diffusion models \citep{liu2023audioldm,evans2024stableaudio} that made high-resolution synthesis feasible. While these systems differ in their cross-modal interaction mechanisms, they predominantly share a core design assumption: corrupting both modalities along a single shared noise schedule, a key limitation our work directly re-examines.

\textbf{Joint Audio-Video Editing. } Instruction-guided audio-video editing requires modifying specific target attributes while preserving unmentioned content intact. To address the resulting data scarcity, JAVEdit \citep{javedit} employs an agent-in-the-loop curation pipeline and benchmark, whereas SpongeBob \citep{spongebob} reformulates editing as self-supervised inpainting on unannotated clips. InstructAV2AV \citep{instructav2av} trains an end-to-end joint editor using gated attention to balance instruction compliance against source preservation. For efficiency, AVControl \citep{avcontrol} deploys lightweight per-modality control adapters, whereas JoVA \citep{jova} and the concurrent MiniMax-H3 \citep{minimaxh3} unify generation and editing within a shared representation space. These existing methods rely heavily on scarce paired editing data to establish basic single-modality competence. In contrast, \model{} derives single-modality capabilities from abundant unimodal corpora, dedicating its paired supervision exclusively to cross-modal coupling.

\textbf{Non-uniform Noise Schedules. } Diffusion models typically apply a single noise level across an entire sample, but two distinct lines of work depart from this assumption. The first varies noise levels across \emph{positions} within a single modality, using independent per-frame noise \citep{chen2024diffusionforcing}, sliding temporal windows \citep{ruhe2024rollingdiffusion,kim2024fifo}, or monotonic token schedules \citep{wu2024arDiffusion}. The second varies noise across \emph{modalities}: UniDiffuser \citep{bao2023unidiffuser} samples independent noise for image and text, while MoNL \citep{monl2024} varies noise along temporal and modal axes to enable arbitrary conditional generation. Extending this to audio and video, MOVA \citep{mova} draws the two timesteps independently with phase-wise shift parameter calibration so no fixed noise-level relation is enforced. While this covers more of the joint noise space, uncoupled sampling breaks the deterministic link between noise pairs and the step counter, leaving the model without structured conditioning during inference. Unison \citep{unison} decouples the timesteps during training via a curriculum from synchronous warmup to full independence, so the leading modality changes freely between samples. Orthogonally, rectified-flow models scale global schedules by sequence length \citep{esser2024scaling}, acknowledging density disparities while still enforcing a single trajectory per sample. In contrast, \model{} assigns dedicated schedules per modality anchored to a shared step counter with the same asymmetric noise-level trajectories at training and sampling. Trajectories remain phase-locked as their magnitudes diverge, keeping the denser visual stream at high noise long after the acoustic stream converges. Finally, two clean-branch subtasks explicitly supervise boundary states where one stream remains clean, directly enabling single-modality editing and post-inference branch refinement.

\section{Method}
\label{sec:method}

\subsection{Preliminary}
\label{sec:formulation}

\textbf{Problem Formulation.} We operate in the latent spaces of pretrained video and audio autoencoders. A sample is defined as a tuple $(c, \vz_0, \az_0, \refv, \refa)$ pairing a text condition $c$ with clean target latents $(\vz_0, \az_0)$ alongside source reference latents $(\refv, \refa)$, which are supplied during editing and set to $\varnothing$ for generation from scratch. A specific task is determined by which references are present and which branches are supervised via binary masks $M_v, M_a \in \{0,1\}$. Joint generation and joint editing both set $M_v=M_a=1$, differing only in whether non-null references are provided. Conversely, cross-modal and single-modality tasks supervise one branch while setting the opposing mask to zero to hold that stream clean. Consequently, generation and editing are unified under a single formulation, distinguished solely by their conditioning inputs and supervision masks.

\textbf{Rectified Flow.} Rectified flow learns a velocity field by regressing the derivative of a
prescribed linear path \citep{liu2023rectified,lipman2023flow}. For a clean latent $\bm{x}_0$ and noise
$\epsilon\sim\mathcal{N}(\bm{0},\bm{I})$, the path is $\bm{x}_t=(1-t)\bm{x}_0+t\,\bm{\epsilon}$ with
$t\sim\mathcal{U}(0,1)$, whose target velocity is
$\bm{u}(\bm{x}_t,t\mid \bm{x}_0,\bm{\epsilon})=\frac{\mathrm{d}\bm{x}_t}{\mathrm{d}t}=\bm{\epsilon}-\bm{x}_0$.
Under this path, the absolute noise level $\sigma$ of $\bm{x}_t$ is numerically interchangeable with $t$, \ie, $\sigma\triangleq t$, hence we use $\sigma$ throughout for clarity. Assigning each modality its own level gives
\begin{equation}
  \vz_{\sigv}=(1-\sigv)\,\vz_0+\sigv\,\bm{\epsilon}_v,
  \qquad
  \az_{\siga}=(1-\siga)\,\az_0+\siga\,\bm{\epsilon}_a,
  \qquad
  \bm{\epsilon}_v,\bm{\epsilon}_a\sim\mathcal{N}(\bm{0},\bm{I}),
  \label{eq:forward}
\end{equation}
with $\bm{\epsilon}_v,\bm{\epsilon}_a$ drawn independently. A velocity network $v_\theta$ is optimized with
\begin{equation}
  \mathcal{L}_{\mathrm{FM}}
  =
  \sum_{i\in\{a,v\}, \xz\in\{\az,\vz\}}\E{\substack{\xz_0,c,\bm{\epsilon}_i,\sigi,\refx}}{%
    w_i \, M_i \big\lVert v_\theta^{(i)}-(\bm{\epsilon}_i-\xz_0)\big\rVert_2^2
  },
  \label{eq:loss}
\vspace{-.5em}
\end{equation}
where $v_\theta^{(v)}$ and $v_\theta^{(a)}$ are the branch velocities predicted from both noised
latents, both noise levels, the text and the references, and $w_v,w_a$ are the branch loss weights.
The expectation is taken over the data distribution of the tuple, the independent noise draws, and
the noise-level pair. 

\subsection{Unified Generation and Editing}
\label{sec:backbone}

\begin{figure*}[t]
\begin{center}
\includegraphics[width=\linewidth]{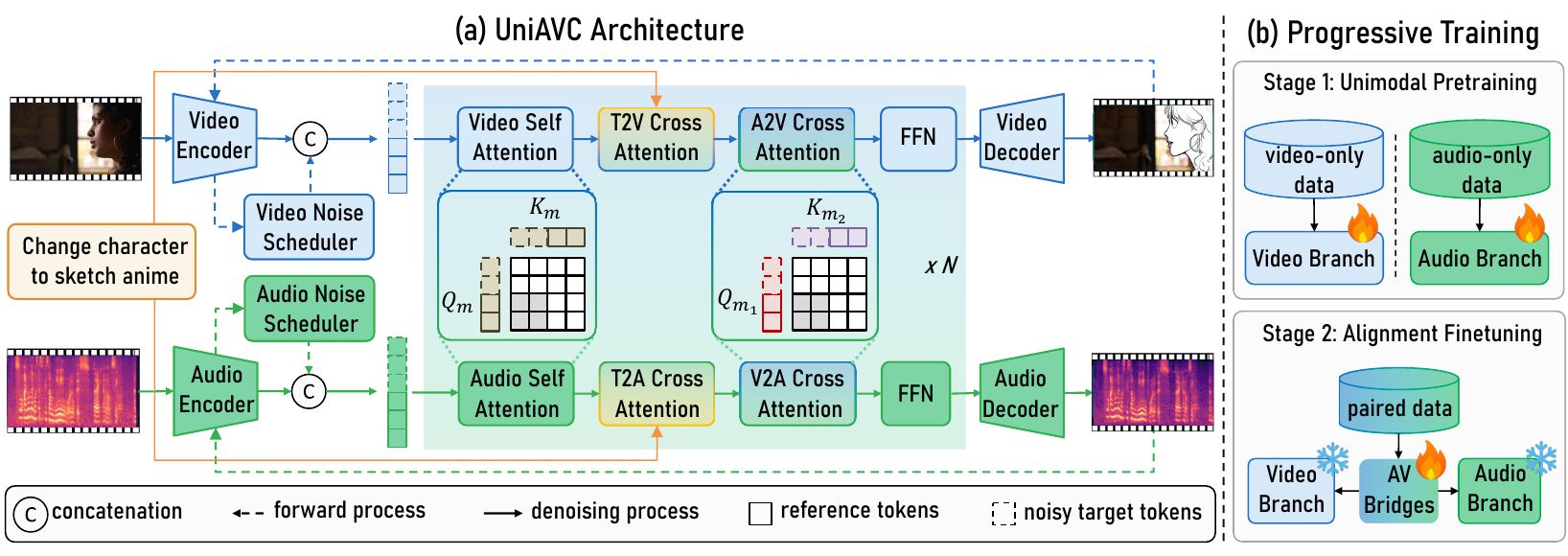}
\end{center}
\caption{Illustration of the \model{} framework. (a) Two transformer stacks process video and audio, intersecting exclusively at per-block cross-attention bridges whose modulation is driven by the partner branch's noise level. Reference latents enter as in-context tokens concatenated to target tokens, with attention masks applied so that reference tokens do not attend to noisy tokens in either self- or cross-attention blocks. (b) Each branch first attains full generative and editing competence on unimodal data in parallel; both branches are subsequently frozen, and the paired corpus is used solely to train the cross-modal bridges.}
\label{fig:arch}
\vspace{-1.5em}
\end{figure*}

\textbf{Backbone. } We build upon LTX-2.3 \citep{ltx2}, where video and audio tokens are processed by separate transformer stacks, each equipped with dedicated self-attention, feed-forward blocks, and adaptive normalization \citep{peebles2023scalable}, while both attend to a shared text embedding. The two branches intersect exclusively at per-block bridges, consisting of an audio-to-video cross-attention layer where video queries attend to audio keys and values, and a symmetric video-to-audio layer with swapped roles. Each bridge operates as a gated residual module whose scale, shift, and gate are derived via adaptive normalization from a noise level. The critical design choice lies in which noise level modulates each path. Crucially, each bridge is modulated by the noise level of the opposing branch, such that the gate regulating information flow from audio into video is a function of $\siga$, while its mirror in the visual branch depends on $\sigv$ (Fig.~\ref{fig:arch}a).


\textbf{In-Context Reference Conditioning. }
Rather than adding dedicated input branches, we adopt in-context conditioning for diffusion transformers \citep{huang2024context} by concatenating reference latents, patchified by the same encoder, onto the target sequence of their corresponding modality. A temporal offset applied to the reference rotary embeddings and a learned token-type embedding separate the two sequence spans. Prior in-context video editing works assign reference tokens the exact positions of their target counterparts, enforcing strict spatiotemporal alignment between input and output \citep{chen2026just,polaczek2025syncloRa}. However, this strict alignment fails when editing instructions require relocating or transforming content. Furthermore, naive concatenation allows reference tokens to attend to noisy target tokens, causing the conditioning signal to drift and degrade as noise increases. We prevent this by applying attention masks that allow target queries to attend to the full sequence while restricting reference queries to read only the uncorrupted reference span. The same isolation extends to cross-modal bridges, where reference queries in one branch can attend to the partner modality's reference span but not its noisy target (Fig.~\ref{fig:arch}a). Because the mask is engaged only when reference latents are present, it applies to editing and is inactive in cross-modal generation, where the conditioning modality occupies the output token slots at $\sigma=0$ rather than the reference slots.

\subsection{Progressive Training Paradigm}
\label{sec:training}

\textbf{Two Stages. } Paired audio-video editing datasets are scarce compared to the abundant unimodal corpora used to pretrain single-modality generators. Training a joint editor directly on limited paired data forces the model to learn source preservation, instruction following, and cross-modal synchronization simultaneously, a multi-task burden known to cause poor convergence \citep{instructav2av}. To resolve this bottleneck, we decompose training into two stages: Unimodal Pretraining and Alignment Finetuning (Fig.~\ref{fig:arch}b). The first stage establishes full generative and editing competence within each independent branch using modality-specific data. The second stage integrates both pretrained branches into the dual-branch backbone and updates only the cross-modal bridges alongside their adaptive normalization parameters. This strategy dedicates scarce paired datasets entirely to learning cross-modal alignment and temporal synchronization, which cannot be acquired from unimodal corpora alone.

\textbf{Conditioning Dropout. } Classifier-free guidance \citep{ho2022classifier} is enabled by randomly dropping the text prompt and reference latents independently during training. This exposes two distinct guidance axes at inference rather than a single scale: the text axis governs instruction compliance, while the reference axis enforces source fidelity, allowing the two signals to be traded off explicitly during sampling. Furthermore, we include a reconstruction subtask where the reference is designated as the target latent under a null-edit instruction. This subtask provides explicit supervision for source preservation, directly counteracting the tendency of instruction-tuned models to alter regions unmentioned by the prompt.

\subsection{Asymmetric Modality-Decoupled Schedulers}
\label{sec:sched}

\textbf{Noise-Level Pair. } Standard diffusion models typically sample a single noise level per trajectory across all modalities \citep{bao2023unidiffuser,monl2024}, imposing the constraint $\sigv=\siga$ on Eq.~\ref{eq:forward}. In contrast, with independently drawn noise vectors $\bm{\epsilon}_v$ and $\bm{\epsilon}_a$, Eq.~\ref{eq:forward} defines a family of corrupted distributions parameterized by the noise-level pair $(\siga,\sigv)$:
\begin{equation}
q_{\siga,\sigv}(\vz,\az \mid \vz_0,\az_0) = \mathcal{N}\big(\vz; (1-\sigv)\vz_0, \sigv^2 \bm{I}\big) \mathcal{N}\big(\az; (1-\siga)\az_0, \siga^2 \bm{I}\big), \quad \siga,\sigv\in[0,1],
\label{eq:family}
\end{equation}
from which each branch regresses its target velocity. Specifically, the video branch targets $\E{\bm{\epsilon}_v\mid \vz_{\sigv},\az_{\siga},c}{\bm{\epsilon}_v-\vz_0}$, while the audio branch targets the analogous vector for its own stream. Because each branch conditions on the partner stream at its respective noise level, three distinct operational regimes emerge. On the diagonal ($\siga=\sigv$), the model observes a partner stream that is informative yet noisy, regressing a velocity that averages over the partner's uncertainty. When $\siga=0$, the partner stream is completely clean, allowing the video branch to treat audio as an uncorrupted constraint, which is essential for cross-modal generation and audio-preserving editing. When $\siga=1$, Eq.~\ref{eq:forward} yields pure noise ($\az_{\siga}=\bm{\epsilon}_a$), making the partner independent of $(\vz_0,\az_0,c)$ so that the branch relies solely on text conditioning. Standard shared schedules visit only the diagonal (Fig.~\ref{fig:c2}), preventing the model from encountering the clean-partner boundary regimes needed for downstream editing and cross-modal generation tasks.

\textbf{Dual-Schedule Shifting. }
A shared synchronization clock does not imply a shared noise level. We maintain a single step index $i\in\{0,\dots,T-1\}$ drawn uniformly per sample while decoupling its mapping to each branch's noise level by warping the base linear grid $\bar\sigma_i=1-i/T$ per modality:
\begin{equation}
  S_k(\sigma) = \frac{k\,\sigma}{1 + (k-1)\,\sigma},
  \quad
  \sigv^{(i)} = \mathcal{R}\big[S_{k_v}\big](\bar\sigma_i),\quad
  \siga^{(i)} = \mathcal{R}\big[S_{k_a}\big](\bar\sigma_i),
  \label{eq:shift}
\end{equation}
where $k_v$ and $k_a$ are per-branch shift strengths and $\mathcal{R}$ rescales the warped grid to a shared terminal noise level. Because $S_k$ is an increasing bijection on $[0,1]$ and $\mathcal{R}$ preserves terminal boundaries, both schedules span identical endpoints. Substituting Eq.~\ref{eq:shift} into Eq.~\ref{eq:family} yields a dual schedule $i\mapsto(\siga^{(i)},\sigv^{(i)})$ that departs from the diagonal throughout its interior. Because $S_k$ is strictly increasing with respect to $k$, setting $k_v>k_a$ ensures $\siga^{(i)}<\sigv^{(i)}$ at every interior step, positioning the schedule strictly below the diagonal (Fig.~\ref{fig:c2}). Audio thus resolves ahead of video, allowing cross-modal bridges to pass a progressively cleaner acoustic representation into the video branch, establishing a natural temporal ordering from two warped schedules on a single index rather than manual staging.

\textbf{Clean-Branch Subtasks. } Because continuous schedules rarely reach the boundary faces where $\sigma=0$ for an individual branch, we explicitly supervise these boundary conditions to support cross-modal generation and single-modality editing. Specifically, we assign a small fraction of training samples to subtasks where one stream is pinned to its uncorrupted latent and its loss is masked, such as audio-conditioned video generation ($\siga=0$, $\bm{M}_a=0$ in Eq.~\ref{eq:loss}) or its visual counterpart. Masking the loss reflects that the uncorrupted stream serves as a fixed conditioning signal rather than a regression target, which guarantees that single-modality edits leave the non-target stream untouched (Sec.~\ref{sec:selective}). Concurrent work \citep{spongebob} adopts similar modes by switching between fixed discrete configurations; in contrast, our unified step index dynamically positions each denoising step along distinct off-diagonal pairs within a continuous noise-level space.

\subsection{Selective Regeneration}
\label{sec:selective}

Dual-schedule training expands the domain of observed noise-level pairs, directly enabling two specialized inference modes. By holding one branch at $\sigma=0$ while denoising the other, these modes remain in-distribution, as clean-branch subtasks explicitly supervise these boundary regimes.

\textbf{Intent-Aware Editing. } Users frequently require targeted edits that modify one modality while preserving the other, such as restyling video while retaining original dialogue. Single-modality editors treat the preserved stream as static context uncoupled from the output, causing the edited result to drift out of alignment \citep{jova,avcontrol,spongebob}. In contrast, we pin the preserved branch at $\siga=0$ as a reference latent. The visual branch thus conditions directly on clean latents $\az_0$ rather than a corrupted trajectory $\az_{\siga}$, referencing acoustic timing, onsets, and lip motion at every step. Because no denoising steps are applied to the pinned branch, the sampler returns the exact source latent rather than an approximated resynthesis, guaranteeing exact preservation by construction. Both editing directions operate without task-specific adapters, whereas evaluating this setup on a single-schedule baseline constitutes an out-of-distribution extrapolation.

\textbf{Self-Refinement. } When a joint generation yields only one satisfactory modality, users can retain the accepted stream and regenerate the other rather than resampling the entire pair. For instance, to refine a generated audio latent $\hat{\az}_0$ while keeping the video branch fixed, a second denoising trajectory is executed after the initial pass completes. Unlike two-stage inference guidance \citep{spongebob}, which adjusts conditioning within a single trajectory, self-refinement re-noises $\hat{\az}_0$ to a noise level $\sigma_a=\lambda$ and denoises along a rescaled schedule $\lambda\siga^{(i)}$ while holding the visual branch clean at $\sigv=0$. This mechanism mirrors the pinning and in-context conditioning of intent-aware editing, differing only in the source of the pinned stream, which uses the first-pass generated latent rather than user-supplied input.

\section{Experiments}
\label{sec:exp}


\subsection{Experimental Setup}
\label{sec:setup}

\begin{wrapfigure}[12]{r}{0.365\linewidth}
\setlength{\columnsep}{3pt}
\vspace{-12pt}
\centering
\makeatletter\def\@captype{table}\makeatother
\vspace{-1.2em}
\caption{Sample budget by stage.}
\label{tab:budget}
\begin{center}
\small
\setlength{\tabcolsep}{3.8pt}
\begin{tabular}{llcc}
\toprule
Stage & Data & \multicolumn{1}{c}{Corpus} & \multicolumn{1}{c}{Seen} \\
\midrule
\multirow{3}{*}{Pretrain (V)}
  & gen & $14.5$M & $33.4$M \\
  & edit & $0.8$M & $0.8$M \\
  & \emph{total} & $\mathbf{15.3}$\textbf{M} & $\mathbf{34.3}$\textbf{M} \\
\midrule
\multirow{3}{*}{Pretrain (A)}
  & gen & $14.0$M & $27.4$M \\
  & edit & $0.9$M & $27.4$M \\
  & \emph{total} & $\mathbf{14.9}$\textbf{M} & $\mathbf{54.9}$\textbf{M} \\
\midrule
\multirow{3}{*}{Align (AV)}
  & gen & $0.46$M & $0.31$M \\
  & edit & $0.28$M & $0.36$M \\
  & \emph{total} & $\mathbf{0.74}$\textbf{M} & $\mathbf{0.67}$\textbf{M} \\
\bottomrule
\end{tabular}
\end{center}
\end{wrapfigure}
\textbf{Implementation Details. }
We initialize each branch from the pretrained weights of LTX-2.3 \citep{ltx2} and train it in the
two stages of Sec.~\ref{sec:training}. Each stage runs at $720$p over $121$\,frames at $24$\,fps with
$16$\,kHz mono audio, with AdamW optimizer at a learning rate of $1\times10^{-5}$ and weight decay $0.01$ in
\texttt{bf16}. Table~\ref{tab:budget} gives the sample budget of each training stage, where alignment
finetuning spends under $1\%$ of the unimodal pretraining in total. Schedulers run with $k_v=9$ and $k_a=2$. The text
and reference dropout probabilities are both set to $0.1$. Inference uses $30$ denoising steps, with text and reference CFG scales of $(3.0, 2.0)$ and $(1.5, 3.0)$ for video and audio respectively, and $\lambda=0.5$ when self-refinement is enabled.

\textbf{Datasets. }
Stage 1 trains each branch on unimodal data, and consumes no paired sample. The video branch
draws $34.3$M samples from a $15.3$M pool built from Boogu-Image \citep{boogu}, Ditto \citep{ditto}
and OpenVE-3M \citep{openve}; the audio branch draws $54.9$M samples from a $14.9$M pool built from
AudioSet \citep{audioset}, VGGSound \citep{vggsound}, WavCaps \citep{wavcaps}, Emilia
\citep{emilia}, LibriTTS \citep{libritts}, AudioEdit \citep{audioomni} and the ISCSLP 2026 CoT-TTS
Challenge \citep{iscslp2026cottts}. Stage 2 then finetunes for alignment on $0.67$M paired
samples drawn from a $0.74$M pool built from TAVGBench \citep{tavgbench}, JAVEdit \citep{javedit}
and InsAVE-80K \citep{instructav2av}.
Table~\ref{tab:budget} summarizes both stages.

\textbf{Benchmarks. }
We evaluate editing and generation capabilities of \model{} on separate benchmarks, in which we follow their evaluation protocol for fair comparison against selected baselines unless otherwise specified. For joint editing, we use JAVEditBench \citet{javedit} and the held-out evaluation split of InsAVE-80K \citep{instructav2av}. For joint generation, we employ Verse-Bench \citep{universe1}, while for cross-modal generation, we adopt the test set of VGGSound \citep{vggsound} with protocol from AVTok \citep{pham2026avtok} for assessment.

\textbf{Baselines. }
We select baselines per capability. For joint editing, we compare against AVED \citep{aved}, AVI-Edit \citep{aviedit}, JAVEdit
\citep{javedit}, InstructAV2AV \citep{instructav2av}, MiniMax-H3
\citep{minimaxh3}, and cascaded methods \citep{coherentavedit,javedit}. For joint generation, we compare against Ovi
\citep{ovi}, JavisDiT \citep{javedit}, UniVerse-1 \citep{universe1}, JoVA \citep{jova}, LTX-2.3 \citep{ltx2}, and MiniMax-H3 \citep{minimaxh3}. Finally, we compare against some representative specialists including MMAudio
\citep{cheng2024mmaudio} and VinTAGe \citep{vintage} for video-to-audio; and AVTok \citep{pham2026avtok} and TempoTokens \citep{tempotokens} for
audio-to-video generation.

\subsection{Main Results}
\label{sec:main}

\begin{table*}[t]
\vspace{-.5em}
\caption{Quantitative comparisons for joint audio-video editing on JAVEditBench
and the test split of InsAVE-80K. Column names abbreviate each benchmark's
official metrics for brevity. $\dagger$ denotes results retrieved from \citep{javedit}.
Best results are in \textbf{bold} and second best are \underline{underlined}.}
\vspace{-1.em}
\label{tab:main}
\begin{center}
\resizebox{\textwidth}{!}{%
\scriptsize
\setlength{\tabcolsep}{3pt}
\begin{tabular}{l cccccc | cccccc}
\toprule
\multirow{2}{*}{\textbf{Method}} & \multicolumn{6}{c|}{\textbf{JAVEditBench}} & \multicolumn{6}{c}{\textbf{InsAVE-80K}} \\
\cmidrule(lr){2-7}\cmidrule(lr){8-13}
 & VQ$\uparrow$ & AQ$\uparrow$ & AVS$\uparrow$ & IC$\uparrow$ & VF$\uparrow$ & AVQ$\uparrow$
 & FVD$\downarrow$ & TV-A$\uparrow$ & FAD$\downarrow$ & TA-A$\uparrow$ & AV-A$\uparrow$ & PEAVS$\uparrow$ \\
\midrule
AVED            & $0.0590$ & $1.72$ & $0.1641$ & $2.95$ & $3.87$ & $2.93$
               & $305.41$ & $24.68$ & $2.86$ & $29.89$ & $26.44$ & $3.15$ \\
AVI-Edit        & $\underline{0.0604}$ & $2.34$ & $0.2721$ & $3.49$ & $3.89$ & $\underline{3.86}$
               & $205.34$ & $25.15$ & $5.09$ & $30.07$ & $26.37$ & $2.54$ \\
Cascaded        & $0.0563$ & $2.35$ & $0.2925$ & $3.99$ & $4.08$ & $3.51$
               & $326.63$ & $25.18$ & $8.01$ & $\underline{32.47}$ & $22.67$ & $3.20$ \\
JAVEdit$^\dagger$& $0.0596$ & $\underline{2.42}$ & $\mathbf{0.3688}$ & $4.07$ & $\underline{4.22}$ & $\mathbf{3.88}$
               & -- & -- & -- & -- & -- & -- \\
InstructAV2AV   & $0.0588$ & $1.40$ & $0.1759$ & $2.88$ & $3.93$ & $3.61$
               & $\underline{180.38}$ & $\underline{25.23}$ & $\underline{2.75}$ & $\mathbf{34.96}$ & $\underline{27.72}$ & $3.24$ \\
MiniMax-H3      & $\mathbf{0.0605}$ & $2.42$ & $0.2307$ & $\underline{4.11}$ & $4.17$ & $3.61$
               & $211.24$ & $20.37$ & $3.25$ & $27.32$ & $25.11$ & $\mathbf{3.75}$ \\
\ours \model{} (ours) & $0.0599$ & $\mathbf{2.59}$ & $\underline{0.2975}$ & $\mathbf{4.42}$ & $\mathbf{4.48}$ & $3.85$
                    & $\mathbf{150.86}$ & $\mathbf{25.66}$ & $\mathbf{1.99}$ & $27.36$ & $\mathbf{28.09}$ & $\underline{3.51}$ \\
\bottomrule
\end{tabular}}
\end{center}
\vspace{-1.5em}
\end{table*}

\begin{table*}[t]
\caption{Quantitative comparisons for joint generation on Verse-Bench (Left), and cross-modal generation on VGGSound test split (Right). $\dagger$ denotes results retrieved from \citep{jova}. Best results are in \textbf{bold} and second best are \underline{underlined}.}
\label{tab:gen}
\begin{center}
\begin{minipage}[t]{0.53\textwidth}
\centering
\scriptsize
\setlength{\tabcolsep}{2.5pt}
\begin{tabular}{l cccccc}
\toprule
\textbf{Method} & MS$\uparrow$ & AS$\uparrow$ & PQ$\uparrow$ & CS$\uparrow$ & LSE-C$\uparrow$ & WER$\downarrow$ \\
\midrule
\multicolumn{7}{l}{\textit{Joint Audio-Video}} \\
UniVerse-1 & $0.20$ & $0.47$ & $5.20$ & $0.16$ & $1.62$ & $0.18$ \\
JavisDiT   & $0.27$ & $0.42$ & $5.49$ & $0.19$ & $0.85$ & $1.00$ \\
Ovi        & $\underline{0.55}$ & $0.47$ & $6.20$ & $0.20$ & $6.61$ & $0.12$ \\
JoVA$^\dagger$       & $\mathbf{0.75}$ & $\mathbf{0.49}$ & $6.23$ & $0.29$ & $6.65$ & $0.11$ \\
LTX-2.3    & $0.37$ & $0.34$ & $6.21$ & $0.26$ & $4.07$ & $\underline{0.09}$ \\
MiniMax-H3 & $0.34$ & $\underline{0.48}$ & $\mathbf{6.87}$ & $\mathbf{0.36}$ & $\mathbf{6.96}$ & $0.13$ \\
\midrule
\ours \model{} (ours) & $0.40$ & $\mathbf{0.49}$ & $\underline{6.26}$ & $\underline{0.30}$ & $\underline{6.72}$ & $\mathbf{0.08}$ \\
\bottomrule
\end{tabular}
\end{minipage}\hfill
\begin{minipage}[t]{0.44\textwidth}
\centering
\scriptsize
\setlength{\tabcolsep}{2.5pt}
\begin{tabular}{l cccc}
\toprule
\textbf{Method} & FVD$\downarrow$ & FAD$\downarrow$ & DeSync$\downarrow$ & IB$\uparrow$ \\
\midrule
\multicolumn{5}{l}{\textit{Audio-to-Video}} \\
TempoTokens & $786.61$ & -- & $1.359$ & $0.132$ \\
AVTok & $\underline{150.26}$ & -- & $\underline{1.317}$ & $\underline{0.143}$ \\
\ours \model{} (ours) & $\mathbf{103.17}$ & -- & $\mathbf{0.773}$ & $\mathbf{0.291}$ \\
\midrule
\multicolumn{5}{l}{\textit{Video-to-Audio}} \\
MMAudio & -- & $\underline{17.09}$ & $\underline{0.813}$ & $\underline{0.291}$ \\
VinTAGe & -- & $80.06$ & $1.294$ & $0.044$ \\
\ours \model{} (ours) & -- & $\mathbf{10.12}$ & $\mathbf{0.650}$ & $\mathbf{0.305}$ \\
\bottomrule
\end{tabular}
\end{minipage}
\end{center}
\vspace{-1.75em}
\end{table*}

\begin{figure*}[t]
\begin{center}
\includegraphics[width=\linewidth]{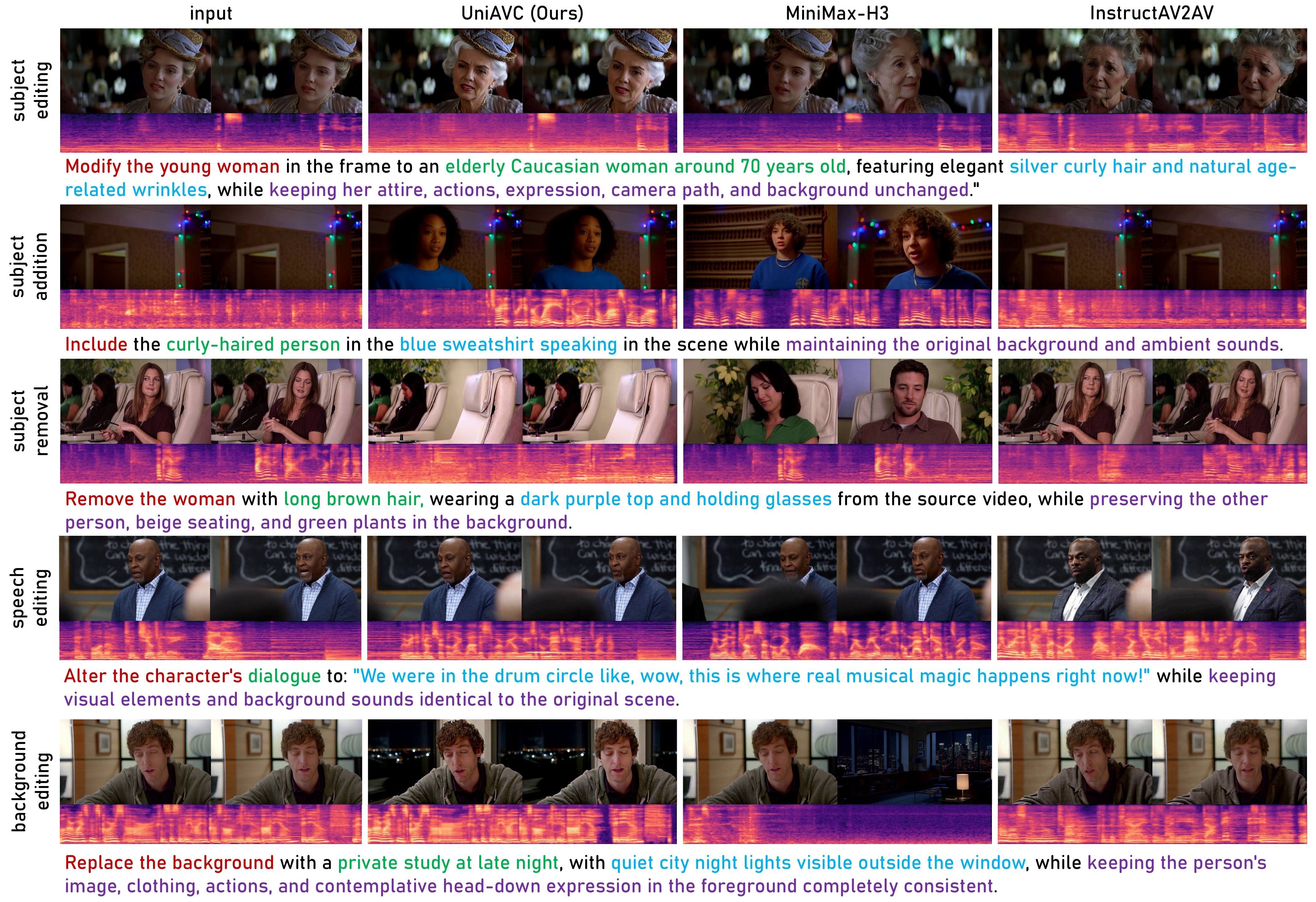}
\end{center}
\caption{Qualitative comparisons for joint editing on JAVEditBench.}
\label{fig:qual}
\vspace{-1.5em}
\end{figure*}

Across both evaluations, \model{} performs competitively against selected baselines across all supported tasks. Specifically, \model{} outperforms prevailing methods on most joint audio-video editing metrics (Tab.~\ref{tab:main}) and simultaneously achieves robust results in both joint and cross-modal generation settings (Tab.~\ref{tab:gen}). Qualitative comparisons in Fig.~\ref{fig:qual} further demonstrate that \model{} maintains precise instruction following, high perceptual fidelity, and robust preservation of character identity and unedited regions, across various tasks, addressing key limitations often observed in the representative InstructAV2AV and MiniMax-H3.

\subsection{Ablation Study}
\label{sec:ablation}

To evaluate the impact of the progressive training recipe, the two inference modes and the
asymmetric dual schedulers proposed in Sec.~\ref{sec:method}, we conduct an ablation study on
JAVEditBench, of which the results are shown in Tab.~\ref{tab:ablation}.

\begin{table*}[t]
\centering
\caption{Ablation study on JAVEditBench.}
\label{tab:ablation}
\resizebox{.8\textwidth}{!}{
\begin{tabular}{@{} ll cccccc @{}}
\toprule
& \textbf{Setting} & VQ$\uparrow$ & AQ$\uparrow$ & AVS$\uparrow$ & IC$\uparrow$ & VF$\uparrow$ & AVQ$\uparrow$ \\
\midrule
\multirow{2}{*}{\textit{Training paradigm}}
 & Single-stage & $0.0584$ & $2.27$ & $0.2446$ & $4.21$ & $4.24$ & $3.53$ \\
 & \cellcolor{gray!15}Progressive & \cellcolor{gray!15}$0.0599$ & \cellcolor{gray!15}$2.59$ & \cellcolor{gray!15}$0.2975$ & \cellcolor{gray!15}$4.42$ & \cellcolor{gray!15}$4.48$ & \multicolumn{1}{>{\columncolor{gray!15}[\tabcolsep][0pt]}c@{}}{$3.85$} \\
\midrule
\multirow{4}{*}{\textit{Inference mode}}
 & Base & $0.0584$ & $2.37$ & $0.2643$ & $4.22$ & $4.31$ & $3.68$ \\
 & Intent-aware & $0.0596$ & $2.53$ & $0.2728$ & $4.31$ & $4.39$ & $3.75$ \\
 & Self-refinement & $0.0592$ & $2.52$ & $0.2643$ & $4.39$ & $4.43$ & $3.75$ \\
 & \cellcolor{gray!15}Full & \cellcolor{gray!15}$0.0599$ & \cellcolor{gray!15}$2.59$ & \cellcolor{gray!15}$0.2975$ & \cellcolor{gray!15}$4.42$ & \cellcolor{gray!15}$4.48$ & \multicolumn{1}{>{\columncolor{gray!15}[\tabcolsep][0pt]}c@{}}{$3.85$} \\
\midrule
\multirow{6}{*}{\shortstack[l]{\textit{Scheduler shift}\\$(k_a,k_v)$}}
 & $(1,1)$ & $0.0589$ & $2.48$ & $0.2545$ & $4.34$ & $4.35$ & $3.67$ \\
 & $(2,2)$ & $0.0592$ & $2.50$ & $0.2607$ & $4.39$ & $4.41$ & $3.80$ \\
 & $(9,9)$ & $0.0592$ & $2.48$ & $0.2608$ & $4.41$ & $4.40$ & $3.77$ \\
 & $(9,2)$ & $0.0591$ & $2.31$ & $0.2520$ & $4.36$ & $4.40$ & $3.58$ \\
 & $(3,12)$ & $0.0592$ & $2.51$ & $0.2618$ & $4.41$ & $4.40$ & $3.80$ \\
 & \cellcolor{gray!15}$(2,9)$ & \cellcolor{gray!15}$0.0599$ & \cellcolor{gray!15}$2.59$ & \cellcolor{gray!15}$0.2975$ & \cellcolor{gray!15}$4.42$ & \cellcolor{gray!15}$4.48$ & \multicolumn{1}{>{\columncolor{gray!15}[\tabcolsep][0pt]}c@{}}{$3.85$} \\
\bottomrule
\end{tabular}%
}
\vspace{-1.5em}
\end{table*}

\textbf{Training Paradigm. } We compare our progressive paradigm against a single-stage baseline. The baseline is initialised from the same LTX-2.3 checkpoint with all parameters trainable and finetuned in one stage over the union of the unimodal and paired corpora, with the same numbers of samples seen as in Tab.~\ref{tab:budget}. A unimodal sample contributes loss only on its own modality while gradients still update the whole model, so the two paradigms consume identical data exposures and differ only in curriculum. Progressive training outperforms single-stage finetuning across all metrics, primarily on cross-modal ones (AVQ, AQ) rather than unimodal quality alone. We attribute this gap to the optimisation conflict between unimodal and cross-modal objectives inside a mixed batch, together with the scarcity of paired supervision that interleaving spends on competence the unimodal corpora already supply.

\textbf{Inference Mode. } Intent-aware editing improves audio quality by conditioning on uncorrupted source latents, whereas self-refinement yields the largest gains in instruction compliance by re-noising a single branch against an accepted counterpart. The two modes operate complementarily, with their combination achieving peak cross-modal synchronization (AVS, AVQ) and overall performance.

\textbf{Scheduler Shift. } Removing schedule asymmetry causes notable drops in cross-modal metrics (AQ, AVQ) with minimal impact on visual quality, demonstrating the limitation of imposing equal noise levels on intrinsically distinctive streams. This holds regardless of shift magnitude, as the symmetric pairs $(2,2)$ and $(9,9)$ both underperform our $(2,9)$ across all metrics, so the gain stems from asymmetry rather than a stronger shared shift. The direction also matters, as reversing the ordering to $(9,2)$, which keeps our shift magnitudes but lets video resolve first, degrades AQ the most. Finally, excessively increasing shift strengths degrades performance across most metrics, showing that schedule warping saturates beyond a moderate shift.

\section{Conclusion}
\label{sec:conclusion}

We present \model{}, a unified framework for audio-video generation and editing, capable of text-conditioned joint generation, bidirectional cross-modal synthesis, and instruction-guided editing of either or both streams without requiring task-specific adapters or secondary checkpoints. This capability rests on two key designs: progressive training establishes unimodal competence from single-modality corpora to reserve paired data for cross-modal bridges, while asymmetric schedulers drive independently warped noise trajectories to access the off-diagonal configurations required for editing and cross-modal generation, complemented by clean-branch subtasks that supervise boundary states. Together, these mechanisms enable intent-aware editing to return preserved modalities unchanged and self-refinement to selectively regenerate flawed modalities. These operational modes differ solely in conditioning and supervision rather than architecture, though coverage remains confined to a 1D trajectory and two boundary faces rather than the full noise-pair space, and cross-modal bridges align representations without repairing unimodal deficits.
\paragraph{AI Use Statement.} Generative AI tools were employed exclusively to polish the manuscript and enhance text clarity and readability. This assistance was strictly limited to linguistic refinement; generative AI played no role in research ideation, data collection or generation, mathematical proofs, experimental design or implementation, data analysis, or the formulation of scientific claims. All AI-assisted revisions were thoroughly reviewed and verified by the authors to guarantee accuracy. The authors retain full responsibility for the entire final manuscript, including all methods, data, analyses, and conclusions.

\bibliography{iclr2027_conference}
\bibliographystyle{iclr2027_conference}

\appendix
\renewcommand\thesection{A\arabic{section}}
\renewcommand\thesubsection{A\arabic{section}.\arabic{subsection}}
\renewcommand\thefigure{A\arabic{figure}}
\renewcommand\thetable{A\arabic{table}}
\renewcommand\theequation{A\arabic{equation}}
\renewcommand\theHfigure{A\arabic{figure}}
\renewcommand\theHtable{A\arabic{table}}
\renewcommand\theHequation{A\arabic{equation}}

\setcounter{section}{0}
\setcounter{figure}{0}
\setcounter{table}{0}
\setcounter{equation}{0}

\section*{Appendix}
{\hypersetup{linkcolor=black}
\startcontents[sections]
\printcontents[sections]{l}{1}{\setcounter{tocdepth}{2}}
}

\clearpage

\section{Additional Implementation Details}
\label{sec:app:data}

\subsection{Training pools and sampling}

Training is divided into two stages based on the target supervision. Unimodal Pretraining optimizes the video and audio branches independently on modality-specific corpora without paired audio-video samples. Alignment Finetuning consumes paired data to establish cross-modal alignment, which unimodal pretraining cannot provide.

\textbf{Unimodal Pretraining (V). } We draw $34.3$M samples from a $15.3$M pool constructed from Boogu-Image \citep{boogu}, Ditto \citep{ditto}, and OpenVE-3M \citep{openve}. Following \citet{loomvideo}, still-image tasks are retained in the mixture to improve frame fidelity and edit diversity. Because image samples consume fewer tokens than $10$s video clips, we assign sampling probabilities of $0.33$ to each image task and $0.17$ to each video task, maintaining an effective balance between spatial fidelity and temporal training dynamics respectively.

\textbf{Unimodal Pretraining (A). } We draw $54.9$M samples from a $14.9$M pool built from AudioSet \citep{audioset}, VGGSound \citep{vggsound}, WavCaps \citep{wavcaps}, Emilia \citep{emilia}, LibriTTS \citep{libritts}, AudioEdit \citep{audioomni}, and the ISCSLP 2026 CoT-TTS Challenge \citep{iscslp2026cottts}, sampling the two audio tasks with equal probability of $0.50$. Because speech samples outnumber environmental sound samples by roughly $5{:}1$ within a single task identifier, we reweight the sampler toward domain-level parity rather than task-level parity.

\textbf{Alignment Finetuning. } We draw $0.67$M paired samples from a $0.74$M pool comprising TAVGBench \citep{tavgbench}, JAVEdit \citep{javedit}, and the training split of InsAVE-80K \citep{instructav2av}, assigning sampling probabilities of $0.30$ to joint generation and $0.70$ to joint editing. This stage operates in a sub-epoch regime, with $0.67$M draws completing roughly $0.90$ of a single pass over the $0.74$M pool.

Tab.~\ref{tab:app:pools} details the per-task breakdown of these dataset sizes and sampling counts.

\begin{table*}[t]
\caption{Detailed decomposition of training pools by stage. ``Seen'' is the number of samples the optimizer consumes during training.}
\label{tab:app:pools}
\begin{center}
\small
\setlength{\tabcolsep}{3pt}
\begin{tabular}{lllrrr}
\toprule
\textbf{Stage} & \textbf{Task} & \textbf{Source} & \textbf{Corpus} & \textbf{Prob.} & \textbf{Seen} \\
\midrule
\multirow{4}{*}{\textit{Unimodal Pretraining} (V)}
 & T2I  & Boogu-Image & $9.5$M & $0.33$ & $16.3$M \\
 & TI2I & Boogu-Image & $3.4$M & $0.33$ & $16.3$M \\
 & T2V  & Proprietary & $1.6$M & $0.17$ & $0.8$M \\
 & TV2V & Ditto, OpenVE-3M & $0.8$M & $0.17$ & $0.8$M \\
\cmidrule(lr){2-6}
 & \multicolumn{2}{l}{\textbf{stage total}} & $\mathbf{15.3}$\textbf{M} & & $\mathbf{34.3}$\textbf{M} \\
\midrule
\multirow{4}{*}{\textit{Unimodal Pretraining} (A)}
 & \multirow{2}{*}{T2A}  & sound: AudioSet, VGGSound, WavCaps & $2.2$M & \multirow{2}{*}{$0.50$} & \multirow{2}{*}{$27.4$M} \\
 &      & speech: Emilia, LibriTTS & $11.8$M & & \\
 & \multirow{2}{*}{TA2A} & AudioEdit & $0.9$M & \multirow{2}{*}{$0.50$} & \multirow{2}{*}{$27.4$M} \\
 &      & CoT-TTS Challenge & $1.0$M & & \\
\cmidrule(lr){2-6}
 & \multicolumn{2}{l}{\textbf{stage total}} & $\mathbf{14.9}$\textbf{M} & & $\mathbf{54.9}$\textbf{M} \\
\midrule
\multirow{2}{*}{\textit{Alignment Finetuning} (AV)}
 & T2AV   & TAVGBench & $0.46$M & $0.30$ & $0.31$M \\
 & TAV2AV & JAVEdit, InsAVE-80K & $0.28$M & $0.70$ & $0.36$M \\
\cmidrule(lr){2-6}
 & \multicolumn{2}{l}{\textbf{stage total}} & $\mathbf{0.74}$\textbf{M} & & $\mathbf{0.67}$\textbf{M} \\
\bottomrule
\end{tabular}
\end{center}
\end{table*}

\subsection{Subtask mixtures}
\label{sec:app:subtasks}

In both training stages, each batch samples a task according to the probabilities in Tab.~\ref{tab:app:pools}. Alignment Finetuning further mixes multiple objectives within each task, allowing a single paired sample to supervise off-diagonal noise configurations in the $(\siga,\sigv)$ space. Specifically, the joint generation task assigns weights of $0.7$ to the standard objective, $0.1$ to reconstruction, and $0.1$ to each clean-branch subtask. The joint editing task assigns weights of $0.8$ to the standard objective and $0.1$ to each self-refinement subtask. 

The reconstruction objective uses the reference as the target alongside an instruction requesting no modification. This supplies the sole supervisory signal for leaving content unaltered, counteracting instruction-following losses that reward target changes without penalizing collateral edits. Clean-branch subtasks explicitly supervise the boundary regime where $\sigma=0$, which is essential for cross-modal generation and single-modality editing, and differ based on whether reference conditioning is present.

\subsection{Evaluation metrics}
\label{sec:app:metrics}

Each benchmark scores a different metric set, and the same abbreviation denotes different quantities
across benchmarks. The metrics are listed below under the benchmark they come from.

\textbf{JAVEditBench.} Six metrics span five dimensions, combining learned perceptual models with a
multimodal judge. \emph{Visual quality} (VQ) is scored by VTSS, the aesthetic model adopted by
Koala-36M \citep{koala36m}. \emph{Audio quality} (AQ) is scored by UTMOSv2 \citep{utmosv2}.
\emph{Audio-visual synchrony} (AVS) is scored by SyncNet as implemented in LatentSync
\citep{latentsync}. \emph{Instruction compliance} (IC), \emph{video fidelity} (VF) and \emph{AV
quality} (AVQ) are judged by Qwen3-Omni \citep{qwen3omni}: whether the edit executes the instruction,
whether unrelated content is preserved, and the overall joint quality of the edited clip.

\textbf{InsAVE-80K.} Six metrics span video, audio and audio-visual properties. \emph{Fr\'echet Video
Distance} (FVD) measures visual realism and spatiotemporal coherence using a VideoMAE \citep{videomae}
backbone. \emph{Fr\'echet Audio Distance} (FAD) is the analogous measure for audio, over VGGish
\citep{vggish} embeddings. \emph{Text-video alignment} (TV-A) and \emph{text-audio alignment} (TA-A)
measure agreement between the instruction and each generated modality, using VideoCLIP-XL
\citep{videoclipxl} and CLAP \citep{wu2023clap}. \emph{Audio-video alignment} (AV-A) measures semantic
coherence between the two generated streams using ImageBind \citep{girdhar2023imagebind}.
\emph{PEAVS} predicts a mean opinion score for audio-visual synchrony \citep{peavs}.

\textbf{VGGSound.} Four metrics are reported under the protocol of \citet{pham2026avtok}. FVD
\citep{fvd} and FAD \citep{fad} measure video and audio quality, DeSync \citep{syncformer} measures
temporal synchronization, and the ImageBind score (IB) \citep{girdhar2023imagebind} measures semantic
alignment. Because the Fr\'echet distances are computed against ground truth, FVD requires generated
video and is reported only for audio-to-video, while FAD requires generated audio and is reported only
for video-to-audio; DeSync and IB are evaluated for both directions. The empty cells in
Tab.~\ref{tab:gen} reflect this, rather than missing measurements.

\textbf{Verse-Bench.} Six metrics are reported following \citet{jova}. \emph{Motion score} (MS) is
computed from RAFT \citep{teed2020raft} optical flow. \emph{Aesthetic score} (AS) averages MANIQA
\citep{yang2022maniqa}, aesthetic-predictor-v2-5 and Musiq \citep{ke2021musiq}. \emph{Production
quality} (PQ) comes from AudioBox-Aesthetics \citep{audiobox}. \emph{Text-audio consistency} (CS) is a
LAION-CLAP \citep{wu2023clap} score. \emph{Lip-sync confidence} (LSE-C) is produced by SyncNet
\citep{syncnet}. \emph{Word error rate} (WER) is computed by transcribing generated audio with
Whisper-large-v3 \citep{whisper}. LSE-C is scored only on human-speech samples, since the metric
requires visible mouth motion.
\section{Additional Experimental Results}
\label{sec:app:exp}

\subsection{Quantitative results}
Tab.~\ref{tab:app:taskwise} further breaks the aggregate of Tab.~\ref{tab:main} down by task family,
where \model{} attains the best or second best score in the majority of cells.

\subsection{Qualitative results}

\begin{figure*}[t]
\begin{center}
\includegraphics[width=0.7\linewidth]{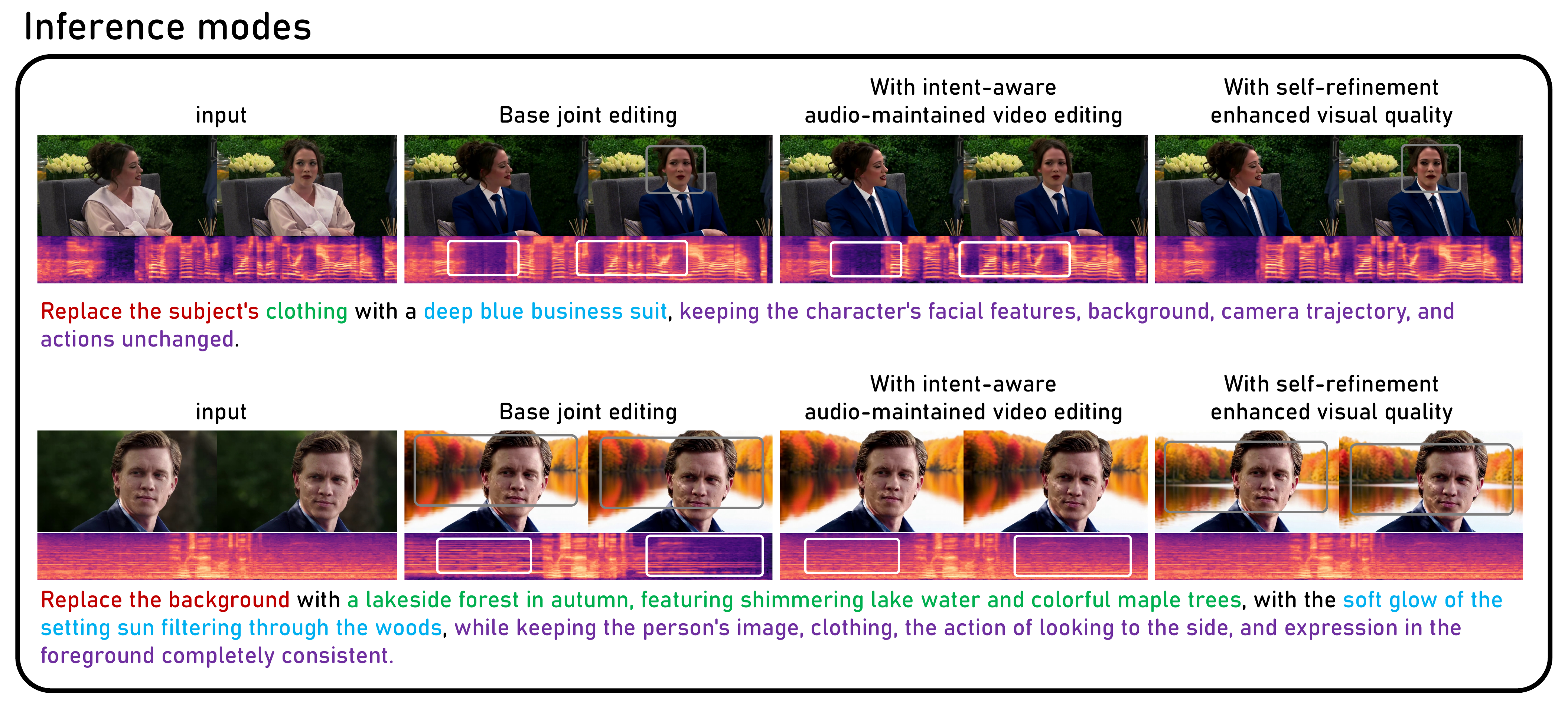}
\end{center}
\caption{Inference modes: intent-aware editing preserves the pinned stream unchanged, while self-refinement improves one branch against the accepted other.}
\vspace{-0.5cm}
\label{fig:app:modes}
\end{figure*}

\begin{figure*}[t]
\begin{center}
\includegraphics[width=.95\linewidth]{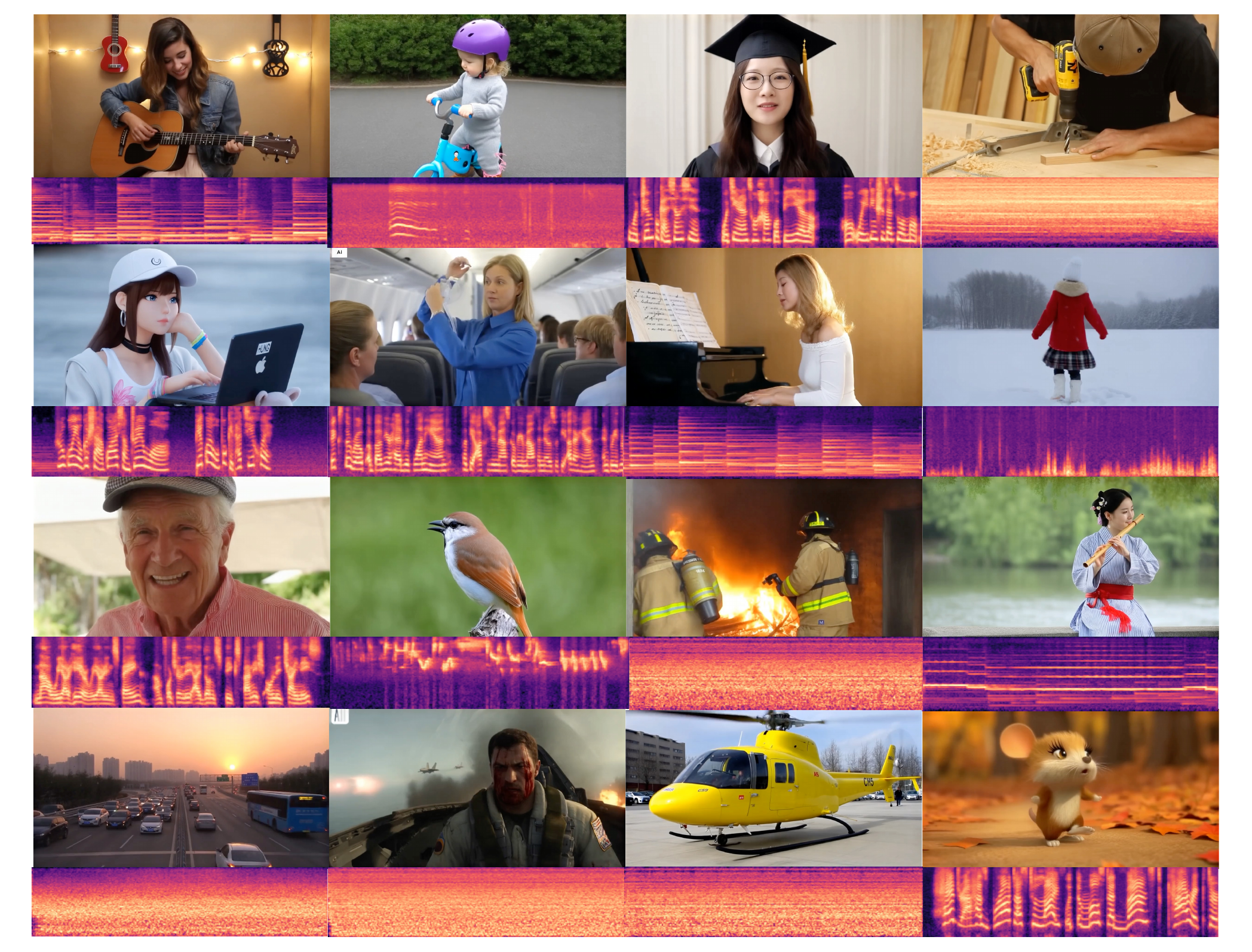}
\end{center}
\caption{Qualitative results for joint audio-video generation on VerseBench. Captions are omitted for brevity.}
\label{fig:app:jointgen}
\end{figure*}

\begin{figure*}[t]
\begin{center}
\includegraphics[width=\linewidth]{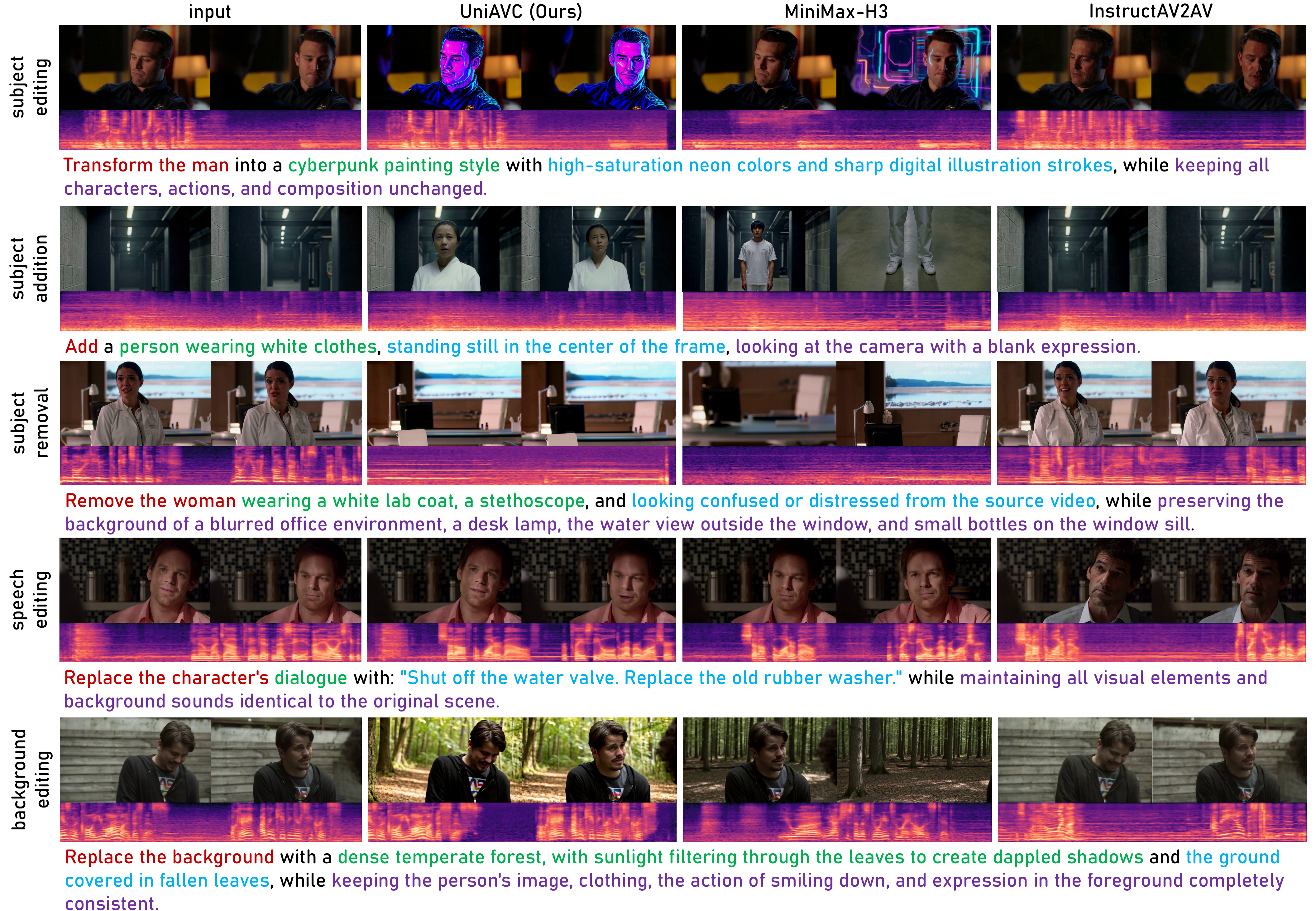}
\end{center}
\caption{Additional qualitative comparisons for joint editing on JAVEditBench.}
\label{fig:app:compare}
\end{figure*}

\begin{figure*}[t]
\centering
\includegraphics[width=0.70\linewidth]{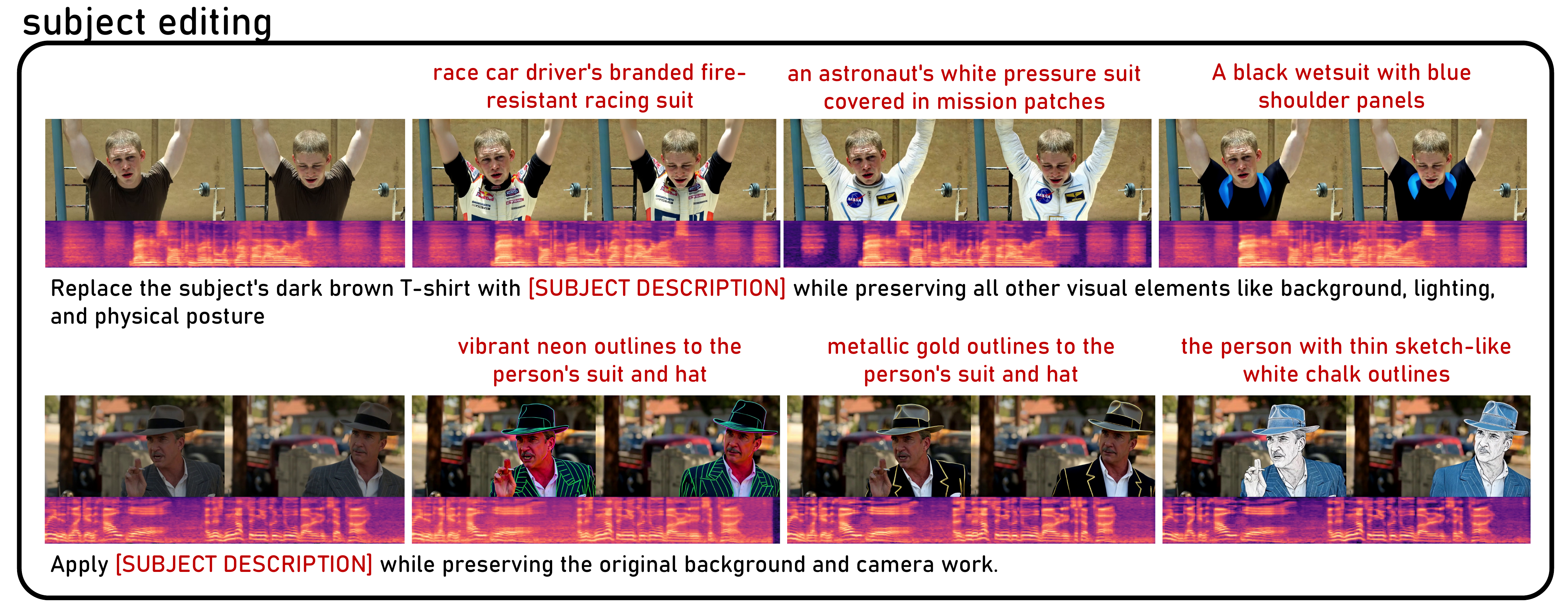} \\[2pt]
\includegraphics[width=0.70\linewidth]{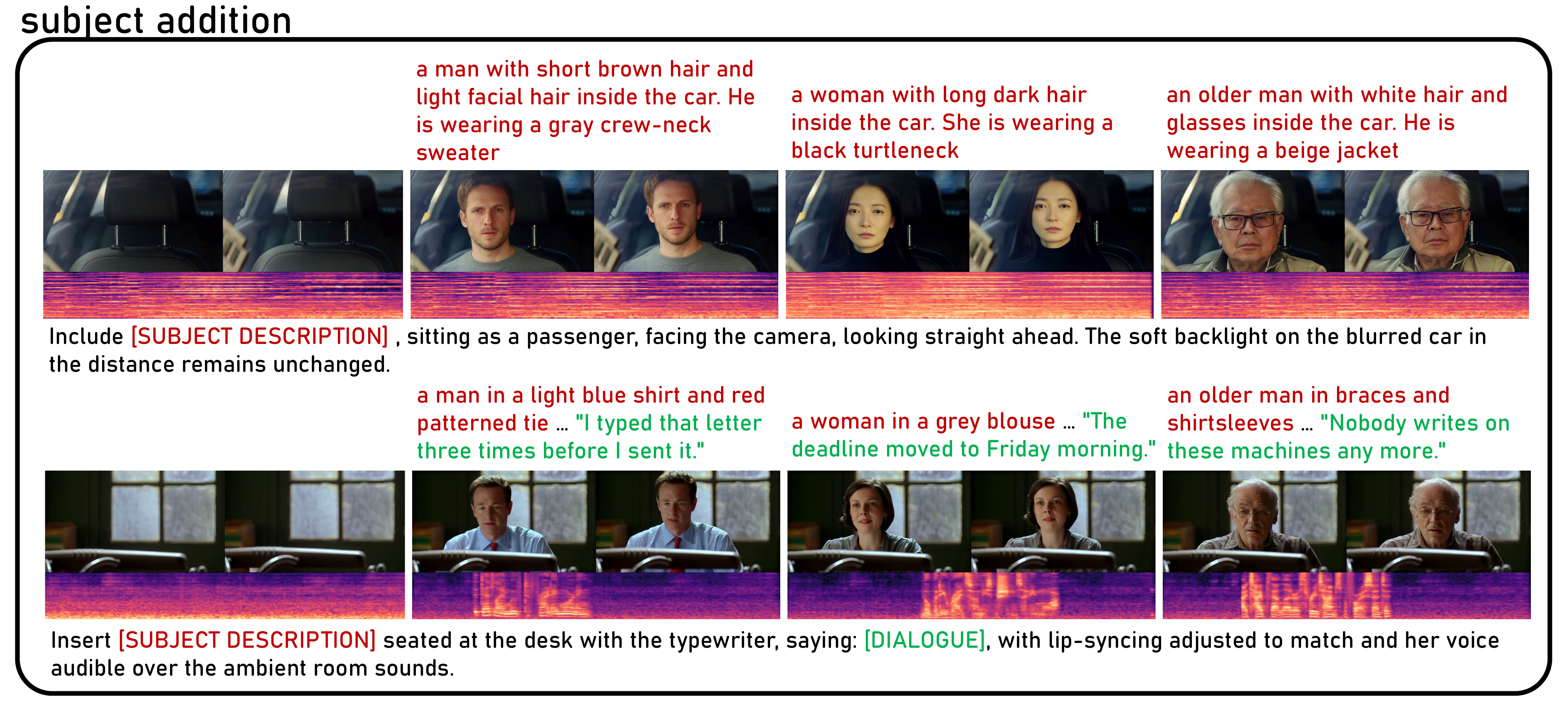} \\[2pt]
\includegraphics[width=0.70\linewidth]{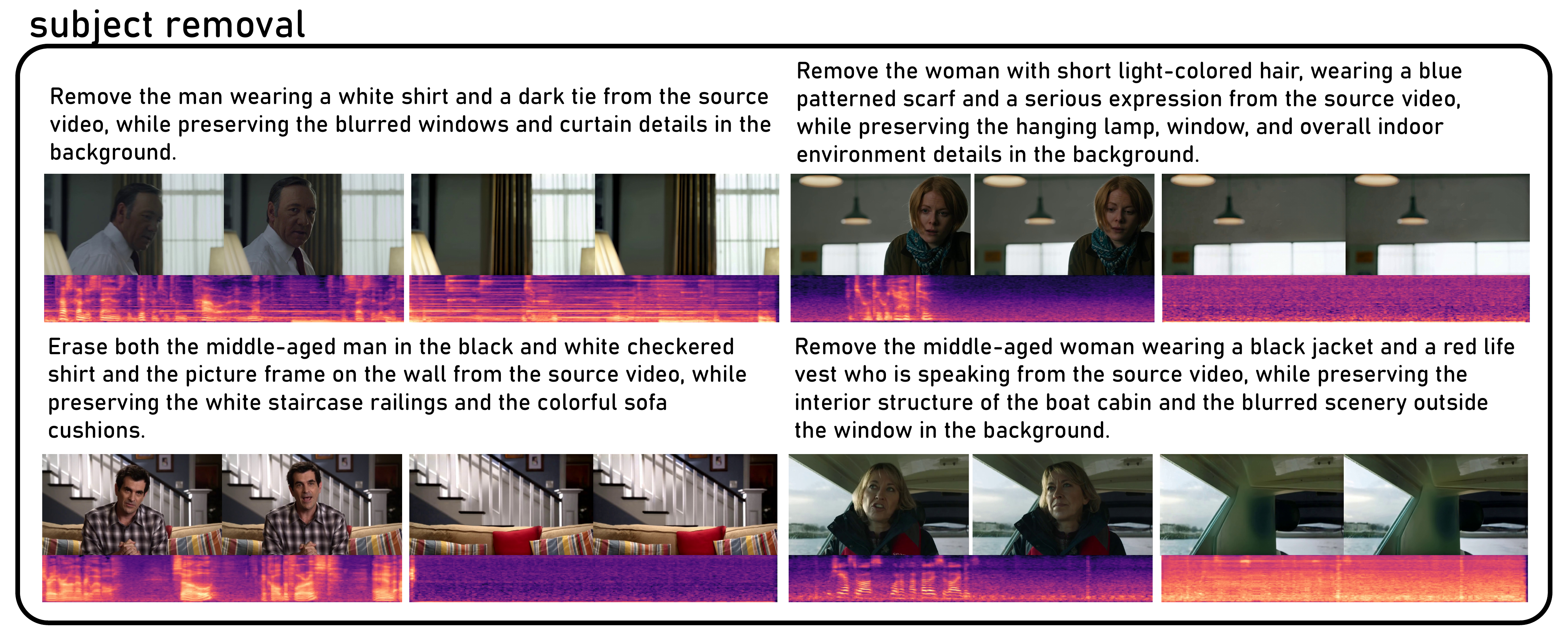} \\[2pt]
\includegraphics[width=0.70\linewidth]{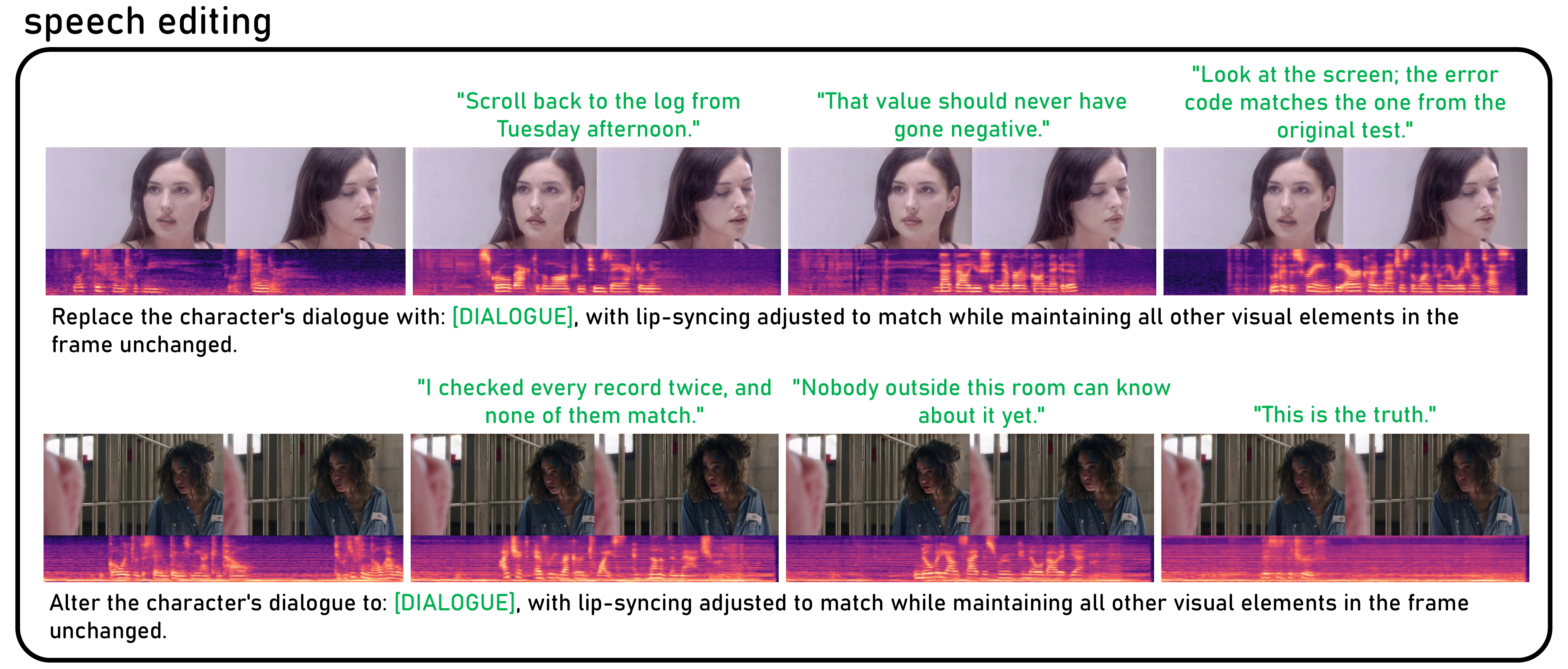} \\[2pt]
\includegraphics[width=0.70\linewidth]{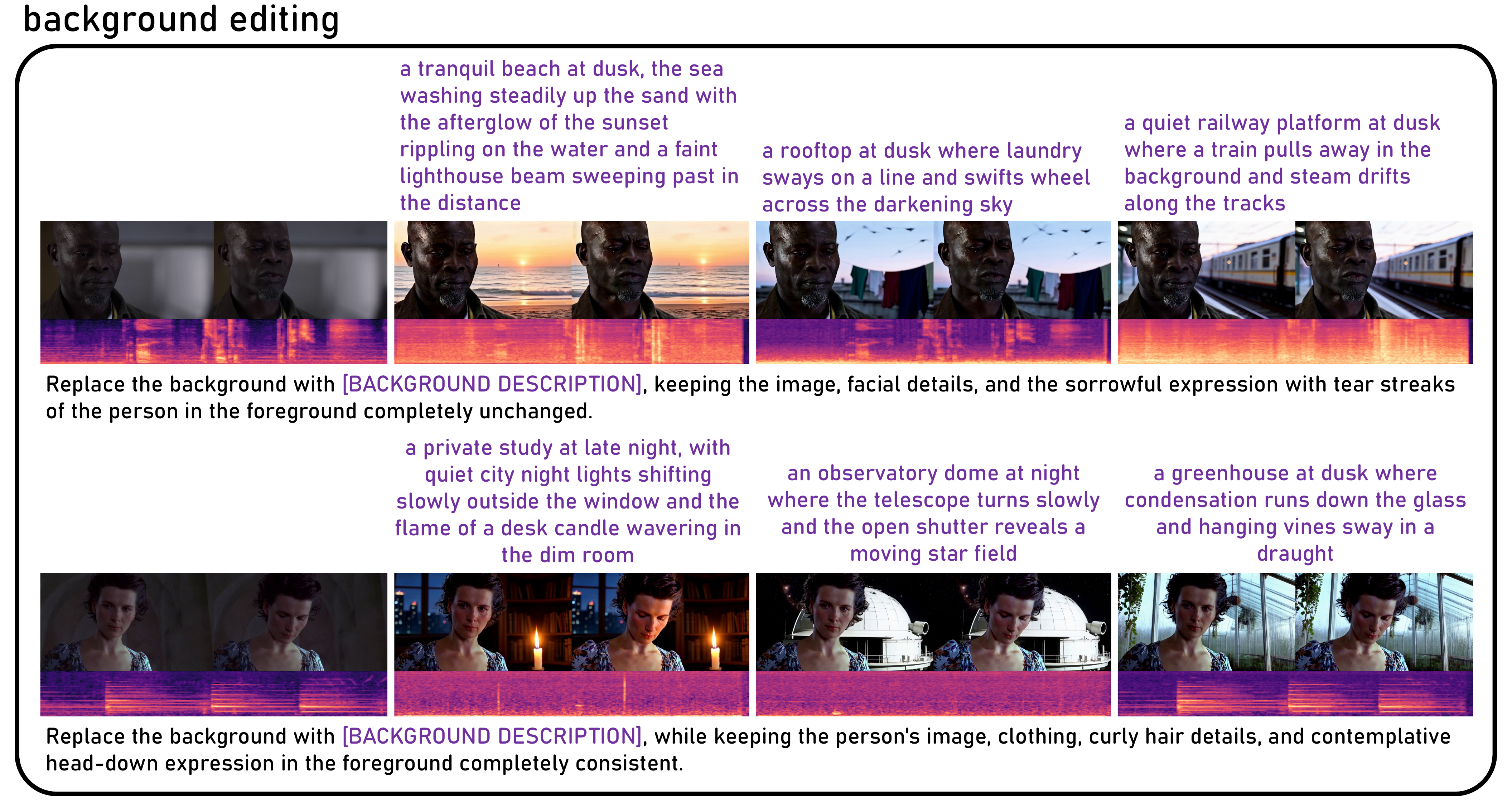}
\caption{Audio-video editing versatility of \model{} across five task families.}
\label{fig:app:editing}
\end{figure*}

Fig.~\ref{fig:app:jointgen} illustrates joint audio-video generation on Verse-Bench, where both streams are generated simultaneously from text prompts alone to yield synchronized soundtracks aligned with the visual scene. Figs.~\ref{fig:app:compare} and \ref{fig:app:editing} compare \model{} against baseline methods and showcase its editing versatility across five task families on matched inputs, demonstrating accurate instruction-guided editing while leaving unmentioned modalities unchanged. Fig.~\ref{fig:app:modes} illustrates the effects of both inference modes on matched inputs. Intent-aware editing and self-refinement both allow for high-quality editing or enhancing the target modality while losslessly preserving the non-target counterpart and maintaining their alignment.

\subsection{Ablations}
\label{sec:app:ablation}

To isolate the effect of clean-branch supervision under the same asymmetric schedule and elaborate on how to determine when self-refinement is applied together with its trade-off, we conduct additional ablation experiments on JAVEditBench, of which the results are shown in Tab.~\ref{tab:app:extra}.

\textbf{Clean-Branch Subtasks.} We conduct another training for the model without clean-branch subtasks while holding all other settings fixed. Since these boundary-state samples are what place intent-aware editing and self-refinement inside the training distribution, neither mode is available to this established variant (3). We compare it against our Base variant (2) with clean-branch subtasks included during training but using identical basic inference mode. AVS and AVQ show significant drops under Base mode, indicating that clean-branch subtasks implicitly help strengthen cross-modal alignment learning without affecting unimodal qualities (VQ, AQ). The Full variant (1), with both inference modes enabled, marks the ceiling that removing the subtasks places out of reach.

\textbf{Self-Refinement Decision \& Trade-off.} Self-refinement requires deciding which stream is satisfactory to preserve and which necessitates further retouching. In practice, we leverage human inspection to subjectively determine this, mirroring users' native creation workflow. However, we conduct an additional experiment and empirically find that Qwen3-Omni can be adopted for reliable judgement, offering an automatic alternative for users when desired. We follow the prompting procedure from JAVEdit~\citep{javedit} for Qwen3-Omni to make the decision, yielding results on par with those of human inspection. This also aligns with JAVEdit's finding of strong positive Spearman correlations with human preference across the judge-based metrics. However, regardless of inspector choices, we note that self-refinement inevitably adds a cost of one extra denoising trajectory, equivalent to a 50\% increase in total NFE at the default $\lambda=0.5$, trading efficiency for enhanced quality.

\subsection{Limitations}
\model{} incurs notable computational overhead during both training and inference. The two-stage recipe requires substantial resources, consuming large-scale unimodal data during pretraining before alignment finetuning begins. Inference latency also precludes real-time deployment because every request requires full denoising trajectories across both streams, with self-refinement adding a second pass over the target modality. Furthermore, performance is constrained by the complexity of instruction prompts in the training set. Because paired corpora contain few compositional requests specifying multiple concurrent edits, the model may execute multi-goal instructions imperfectly. Lastly, the scarcity of paired audio-video data that motivates our progressive paradigm also hinders a paired-data scaling study, which would more directly substantiate its training efficiency as paired corpora grow. Future work will focus on reducing staged training costs, accelerating inference trajectories, expanding instruction mixtures to improve compositional editing, and constructing large-scale paired datasets for scaling study.

\begin{table*}[t]
\caption{Per-task results on JAVEditBench. Best per column in \textbf{bold} and second best \underline{underlined}.}
\label{tab:app:taskwise}
\begin{center}
\small
\begin{tabular}{ll cccccc}
\toprule
\multirow{2}{*}{\textbf{Metric}} & \multirow{2}{*}{\textbf{Method}} & \multirow{2}{*}{\textbf{Overall}} & \textbf{Subject} & \textbf{Background} & \textbf{Speech} & \textbf{Subject} & \textbf{Subject} \\
 & & & \textbf{Editing} & \textbf{Editing} & \textbf{Editing} & \textbf{Addition} & \textbf{Removal} \\
\midrule
\multirow{6}{*}{VQ$\uparrow$} & AVED & $0.0590$ & $0.0603$ & $0.0612$ & $0.0603$ & $0.0551$ & $0.0562$ \\
& AVI-Edit & $\underline{0.0604}$ & $\underline{0.0625}$ & $\mathbf{0.0650}$ & $0.0599$ & $0.0492$ & $\mathbf{0.0638}$ \\
& JAVEdit & $0.0596$ & $0.0609$ & $\underline{0.0636}$ & $\underline{0.0625}$ & $\mathbf{0.0633}$ & $0.0447$ \\
& InstructAV2AV & $0.0588$ & $0.0620$ & $0.0618$ & $0.0619$ & $0.0417$ & $\underline{0.0628}$ \\
& MiniMax-H3 & $\mathbf{0.0605}$ & $0.0609$ & $0.0618$ & $0.0614$ & $0.0618$ & $0.0557$ \\
& \multicolumn{1}{>{\columncolor{gray!15}[\tabcolsep][\tabcolsep]}l}{\model{} (ours)} & \multicolumn{1}{>{\columncolor{gray!15}[\tabcolsep][\tabcolsep]}c}{$0.0599$} & \multicolumn{1}{>{\columncolor{gray!15}[\tabcolsep][\tabcolsep]}c}{$\mathbf{0.0637}$} & \multicolumn{1}{>{\columncolor{gray!15}[\tabcolsep][\tabcolsep]}c}{$0.0634$} & \multicolumn{1}{>{\columncolor{gray!15}[\tabcolsep][\tabcolsep]}c}{$\mathbf{0.0631}$} & \multicolumn{1}{>{\columncolor{gray!15}[\tabcolsep][\tabcolsep]}c}{$\underline{0.0619}$} & \multicolumn{1}{>{\columncolor{gray!15}[\tabcolsep][\tabcolsep]}c}{$0.0423$} \\
\midrule
\multirow{6}{*}{AQ$\uparrow$} & AVED & $1.72$ & $1.71$ & $1.49$ & $1.96$ & $1.62$ & $1.75$ \\
& AVI-Edit & $2.34$ & $2.43$ & $2.28$ & $2.33$ & $2.17$ & $\mathbf{2.44}$ \\
& JAVEdit & $\underline{2.42}$ & $\underline{2.48}$ & $2.23$ & $2.62$ & $\underline{2.48}$ & $2.24$ \\
& InstructAV2AV & $1.40$ & $1.24$ & $1.35$ & $2.05$ & $1.13$ & $1.20$ \\
& MiniMax-H3 & $\underline{2.42}$ & $2.32$ & $\underline{2.31}$ & $\underline{2.69}$ & $\mathbf{2.57}$ & $2.25$ \\
& \multicolumn{1}{>{\columncolor{gray!15}[\tabcolsep][\tabcolsep]}l}{\model{} (ours)} & \multicolumn{1}{>{\columncolor{gray!15}[\tabcolsep][\tabcolsep]}c}{$\mathbf{2.59}$} & \multicolumn{1}{>{\columncolor{gray!15}[\tabcolsep][\tabcolsep]}c}{$\mathbf{2.59}$} & \multicolumn{1}{>{\columncolor{gray!15}[\tabcolsep][\tabcolsep]}c}{$\mathbf{2.52}$} & \multicolumn{1}{>{\columncolor{gray!15}[\tabcolsep][\tabcolsep]}c}{$\mathbf{2.82}$} & \multicolumn{1}{>{\columncolor{gray!15}[\tabcolsep][\tabcolsep]}c}{$\mathbf{2.57}$} & \multicolumn{1}{>{\columncolor{gray!15}[\tabcolsep][\tabcolsep]}c}{$\underline{2.40}$} \\
\midrule
\multirow{6}{*}{AVS$\uparrow$} & AVED & $0.1641$ & $0.1333$ & $0.0613$ & $0.1050$ & $\mathbf{0.5459}$ & $0.1367$ \\
& AVI-Edit & $0.2721$ & $\underline{0.3236}$ & $\mathbf{0.3584}$ & $0.2453$ & $0.0365$ & $\mathbf{0.2658}$ \\
& JAVEdit & $\mathbf{0.3688}$ & $\mathbf{0.3779}$ & $\underline{0.3097}$ & $\mathbf{0.4569}$ & $\underline{0.3264}$ & $0.0146$ \\
& InstructAV2AV & $0.1759$ & $0.1681$ & $0.1531$ & $0.2212$ & $0.0293$ & $\underline{0.1661}$ \\
& MiniMax-H3 & $0.2307$ & $0.2300$ & $0.2046$ & $0.2998$ & $0.2111$ & $0.1643$ \\
& \multicolumn{1}{>{\columncolor{gray!15}[\tabcolsep][\tabcolsep]}l}{\model{} (ours)} & \multicolumn{1}{>{\columncolor{gray!15}[\tabcolsep][\tabcolsep]}c}{$\underline{0.2975}$} & \multicolumn{1}{>{\columncolor{gray!15}[\tabcolsep][\tabcolsep]}c}{$0.3203$} & \multicolumn{1}{>{\columncolor{gray!15}[\tabcolsep][\tabcolsep]}c}{$0.2829$} & \multicolumn{1}{>{\columncolor{gray!15}[\tabcolsep][\tabcolsep]}c}{$\underline{0.3229}$} & \multicolumn{1}{>{\columncolor{gray!15}[\tabcolsep][\tabcolsep]}c}{$0.2672$} & \multicolumn{1}{>{\columncolor{gray!15}[\tabcolsep][\tabcolsep]}c}{$0.0579$} \\
\midrule
\multirow{6}{*}{IC$\uparrow$} & AVED & $2.95$ & $2.60$ & $2.87$ & $3.67$ & $2.20$ & $3.50$ \\
& AVI-Edit & $3.49$ & $3.64$ & $3.63$ & $3.30$ & $3.20$ & $3.58$ \\
& JAVEdit & $4.07$ & $\underline{4.22}$ & $3.83$ & $4.33$ & $3.88$ & $\underline{3.96}$ \\
& InstructAV2AV & $2.88$ & $3.00$ & $2.47$ & $4.10$ & $1.50$ & $3.12$ \\
& MiniMax-H3 & $\underline{4.11}$ & $4.08$ & $\underline{4.37}$ & $\underline{4.40}$ & $\underline{4.12}$ & $3.48$ \\
& \multicolumn{1}{>{\columncolor{gray!15}[\tabcolsep][\tabcolsep]}l}{\model{} (ours)} & \multicolumn{1}{>{\columncolor{gray!15}[\tabcolsep][\tabcolsep]}c}{$\mathbf{4.42}$} & \multicolumn{1}{>{\columncolor{gray!15}[\tabcolsep][\tabcolsep]}c}{$\mathbf{4.35}$} & \multicolumn{1}{>{\columncolor{gray!15}[\tabcolsep][\tabcolsep]}c}{$\mathbf{4.47}$} & \multicolumn{1}{>{\columncolor{gray!15}[\tabcolsep][\tabcolsep]}c}{$\mathbf{4.53}$} & \multicolumn{1}{>{\columncolor{gray!15}[\tabcolsep][\tabcolsep]}c}{$\mathbf{4.40}$} & \multicolumn{1}{>{\columncolor{gray!15}[\tabcolsep][\tabcolsep]}c}{$\mathbf{4.36}$} \\
\midrule
\multirow{6}{*}{VF$\uparrow$} & AVED & $3.87$ & $3.92$ & $4.07$ & $4.00$ & $3.48$ & $3.79$ \\
& AVI-Edit & $3.89$ & $4.10$ & $3.93$ & $3.70$ & $3.72$ & $3.88$ \\
& JAVEdit & $\underline{4.22}$ & $\underline{4.30}$ & $4.07$ & $\underline{4.33}$ & $\underline{4.04}$ & $\underline{4.30}$ \\
& InstructAV2AV & $3.93$ & $4.25$ & $4.10$ & $3.80$ & $\underline{4.04}$ & $3.22$ \\
& MiniMax-H3 & $4.17$ & $4.10$ & $\mathbf{4.33}$ & $4.23$ & $3.88$ & $4.29$ \\
& \multicolumn{1}{>{\columncolor{gray!15}[\tabcolsep][\tabcolsep]}l}{\model{} (ours)} & \multicolumn{1}{>{\columncolor{gray!15}[\tabcolsep][\tabcolsep]}c}{$\mathbf{4.48}$} & \multicolumn{1}{>{\columncolor{gray!15}[\tabcolsep][\tabcolsep]}c}{$\mathbf{4.65}$} & \multicolumn{1}{>{\columncolor{gray!15}[\tabcolsep][\tabcolsep]}c}{$\underline{4.23}$} & \multicolumn{1}{>{\columncolor{gray!15}[\tabcolsep][\tabcolsep]}c}{$\mathbf{4.53}$} & \multicolumn{1}{>{\columncolor{gray!15}[\tabcolsep][\tabcolsep]}c}{$\mathbf{4.36}$} & \multicolumn{1}{>{\columncolor{gray!15}[\tabcolsep][\tabcolsep]}c}{$\mathbf{4.56}$} \\
\midrule
\multirow{6}{*}{AVQ$\uparrow$} & AVED & $2.93$ & $3.10$ & $3.13$ & $3.73$ & $1.72$ & $2.68$ \\
& AVI-Edit & $\underline{3.86}$ & $4.03$ & $3.93$ & $3.87$ & $3.40$ & $\mathbf{3.96}$ \\
& JAVEdit & $\mathbf{3.88}$ & $\underline{4.08}$ & $\mathbf{4.07}$ & $3.63$ & $\underline{3.72}$ & $\underline{3.82}$ \\
& InstructAV2AV & $3.61$ & $3.60$ & $3.63$ & $\underline{3.90}$ & $3.28$ & $3.60$ \\
& MiniMax-H3 & $3.61$ & $3.75$ & $3.67$ & $3.83$ & $3.58$ & $3.08$ \\
& \multicolumn{1}{>{\columncolor{gray!15}[\tabcolsep][\tabcolsep]}l}{\model{} (ours)} & \multicolumn{1}{>{\columncolor{gray!15}[\tabcolsep][\tabcolsep]}c}{$3.85$} & \multicolumn{1}{>{\columncolor{gray!15}[\tabcolsep][\tabcolsep]}c}{$\mathbf{4.15}$} & \multicolumn{1}{>{\columncolor{gray!15}[\tabcolsep][\tabcolsep]}c}{$\underline{3.97}$} & \multicolumn{1}{>{\columncolor{gray!15}[\tabcolsep][\tabcolsep]}c}{$\mathbf{4.00}$} & \multicolumn{1}{>{\columncolor{gray!15}[\tabcolsep][\tabcolsep]}c}{$\mathbf{4.16}$} & \multicolumn{1}{>{\columncolor{gray!15}[\tabcolsep][\tabcolsep]}c}{$2.76$} \\
\bottomrule
\end{tabular}
\end{center}
\end{table*}

\begin{table*}[t]
\caption{Additional ablation study on JAVEditBench. CB and SR denote clean-branch subtasks and self-refinement, respectively.}
\vspace{0.25cm}
\label{tab:app:extra}
\centering
\resizebox{.65\linewidth}{!}{%
\begin{tabular}{ccc cccccc}
\toprule
& \textbf{CB} & \textbf{SR} & VQ$\uparrow$ & AQ$\uparrow$ & AVS$\uparrow$ & IC$\uparrow$ & VF$\uparrow$ & AVQ$\uparrow$ \\
\midrule
\cellcolor{gray!15}(1) & \cellcolor{gray!15}\ding{51} & \cellcolor{gray!15}\ding{51} & \cellcolor{gray!15}$0.0599$ & \cellcolor{gray!15}$2.59$ & \cellcolor{gray!15}$0.2975$ & \cellcolor{gray!15}$4.42$ & \cellcolor{gray!15}$4.48$ & \cellcolor{gray!15}$3.85$ \\
(2) & \ding{51} & \ding{55} & $0.0584$ & $2.37$ & $0.2643$ & $4.22$ & $4.31$ & $3.68$ \\
(3) & \ding{55} & \ding{55} & $0.0579$ & $2.36$ & $0.2545$ & $4.23$ & $4.27$ & $3.55$ \\
\midrule
\multicolumn{3}{c}{\cellcolor{gray!15}Human} & \cellcolor{gray!15}$0.0599$ & \cellcolor{gray!15}$2.59$ & \cellcolor{gray!15}$0.2975$ & \cellcolor{gray!15}$4.42$ & \cellcolor{gray!15}$4.48$ & \cellcolor{gray!15}$3.85$ \\
\multicolumn{3}{c}{Qwen3-Omni} & $0.0595$ & $2.50$ & $0.2854$ & $4.24$ & $4.46$ & $3.82$ \\
\bottomrule
\end{tabular}%
}
\end{table*}

\end{document}